\documentclass[journal]{IEEEtran}
\usepackage{multirow}
\usepackage[table,xcdraw]{xcolor}
\usepackage{amssymb}
\usepackage{pifont}
\usepackage{booktabs}

\usepackage{cite}

\ifCLASSINFOpdf
  \usepackage[pdftex]{graphicx}
\else
\fi
\usepackage{amsmath}
\usepackage{algorithmic}

\usepackage{array}

\ifCLASSOPTIONcompsoc
 \usepackage[caption=false,font=normalsize,labelfont=sf,textfont=sf]{subfig}
\else
 \usepackage[caption=false,font=footnotesize]{subfig}
\fi

\begin{document}
%
\title{Universal-to-Specific Framework for Complex Action Recognition}
%
%
%


\author{Peisen Zhao,
Lingxi Xie,
Ya Zhang,
and~Qi Tian 
\thanks{Peisen Zhao and Ya Zhang are with the Cooperative Medianet Innovation Center, Shanghai Jiao Tong University, Shanghai 200240, China (e-mail: \{pszhao, ya\_zhang\}@sjtu.edu.cn). Lingxi Xie and Qi Tian are with Huawei Noah's Ark Lab, Shenzhen, Guangdong 518129, China (e-mail: 198808xc@gmail.com; tian.qi1@huawei.com). Ya Zhang is the corresponding author. This work is supported by the National Key Research and Development Program of China (No. 2019YFB1804304), SHEITC (No. 2018-RGZN-02046), NSFC (No. 61521062), 111 plan (No. B07022), and STCSM (No. 18DZ2270700).}

\thanks{}}

%
%

\markboth{Journal of \LaTeX\ Class Files, October~2019}
{Shell \MakeLowercase{\textit{et al.}}: Bare Demo of IEEEtran.cls for IEEE Journals}
%



\maketitle

\begin{abstract}
Video-based action recognition has recently attracted much attention in the field of computer vision. To solve more complex recognition tasks, it has become necessary to distinguish different levels of interclass variations. Inspired by a common flowchart based on the human decision-making process that first narrows down the probable classes and then applies a ``rethinking'' process for finer-level recognition, we propose an effective universal-to-specific (U2S) framework for complex action recognition.
The U2S framework is composed of three subnetworks: a universal network, a category-specific network, and a mask network.
The universal network first learns universal feature representations. The mask network then generates attention masks for confusing classes through category regularization based on the output of the universal network. The mask is further used to guide the category-specific network for class-specific feature representations.
The entire framework is optimized in an end-to-end manner. Experiments on a variety of benchmark datasets, e.g., the Something-Something, UCF101, and HMDB51 datasets, demonstrate the effectiveness of the U2S framework; i.e., U2S can focus on discriminative spatiotemporal regions for confusing categories. We further visualize the relationship between different classes, showing that U2S indeed improves the discriminability of learned features. Moreover, the proposed U2S model is a general framework and may adopt any base recognition network.
\end{abstract}


\ifCLASSOPTIONpeerreview
\begin{center} \bfseries EDICS Category: 3-BBND \end{center}
\fi
%
\IEEEpeerreviewmaketitle

\section{Introduction}
%
%
%
%
\IEEEPARstart{V}{ideo-based} action recognition has recently become an important research direction in computer vision. Early studies~\cite{kuehne2011hmdb,soomro2012ucf101,wang2013improvedtrajec,simonyan2014two} started with classifying simple motion states such as ``{\em jumping}'' and ``{\em running}'', which can be easily achieved by directly extracting features from key frames~\cite{wang2017temporal}. Recently, more challenging recognition tasks have been proposed, which require distinguishing fine-grained interclass variations. For example, in the Something-Something dataset~\cite{goyal2017something}, the two action categories ``{\em bending something so that it deforms}'' and ``{\em bending something until it breaks}'' share the same action, {\em bending}, but differ in their consequences. In other words, the difference between a pair of action categories can be either subtle ({\em e.g.}, the difference between two ``bending'' actions) or considerable ({\em e.g.}, the difference between jumping and running). This type of recognition scenario requires the simultaneous representation of both coarse-grained and fine-grained concepts, which is still challenging in the computer vision field.

\begin{figure}[t]
\begin{center}
  \includegraphics[width=0.49\linewidth]{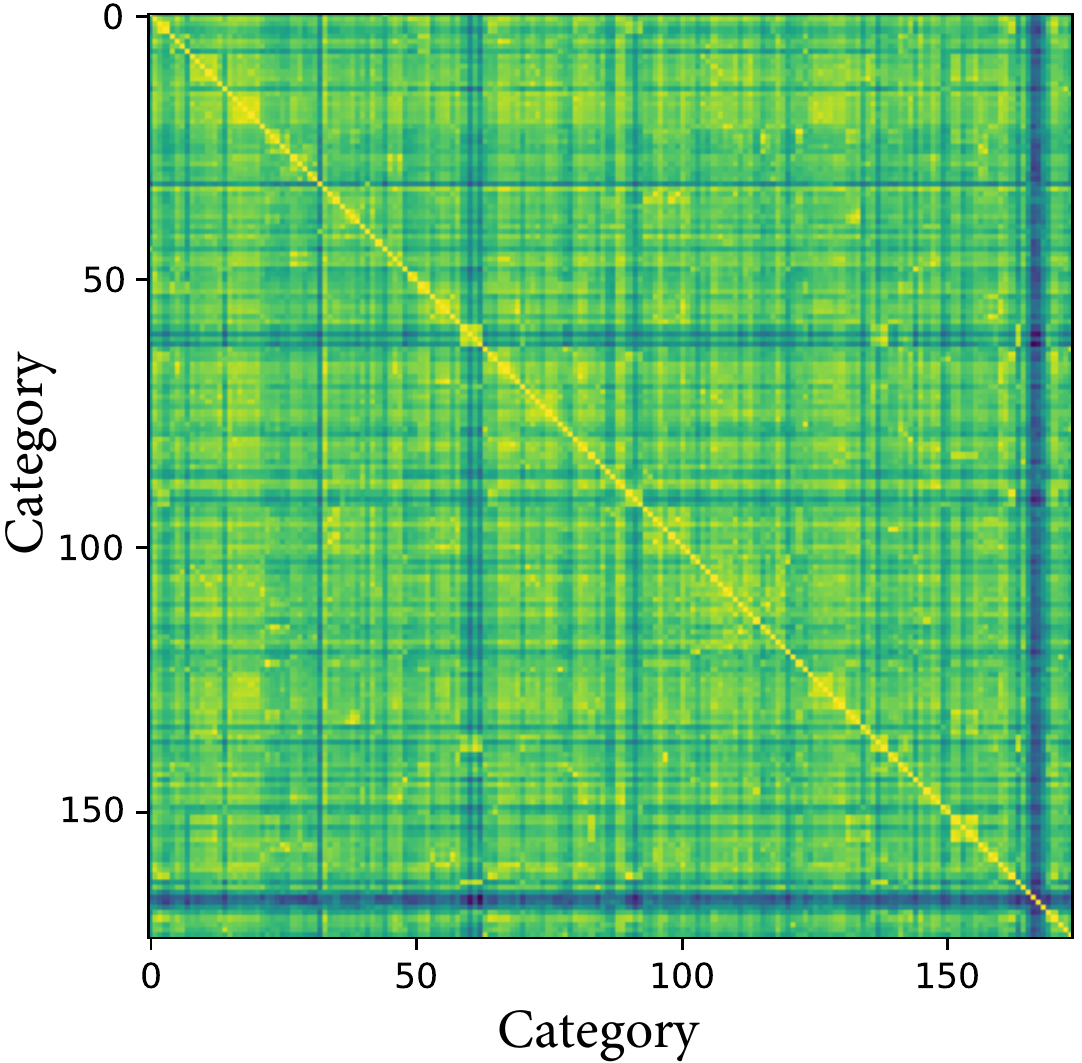}
  \includegraphics[width=0.49\linewidth]{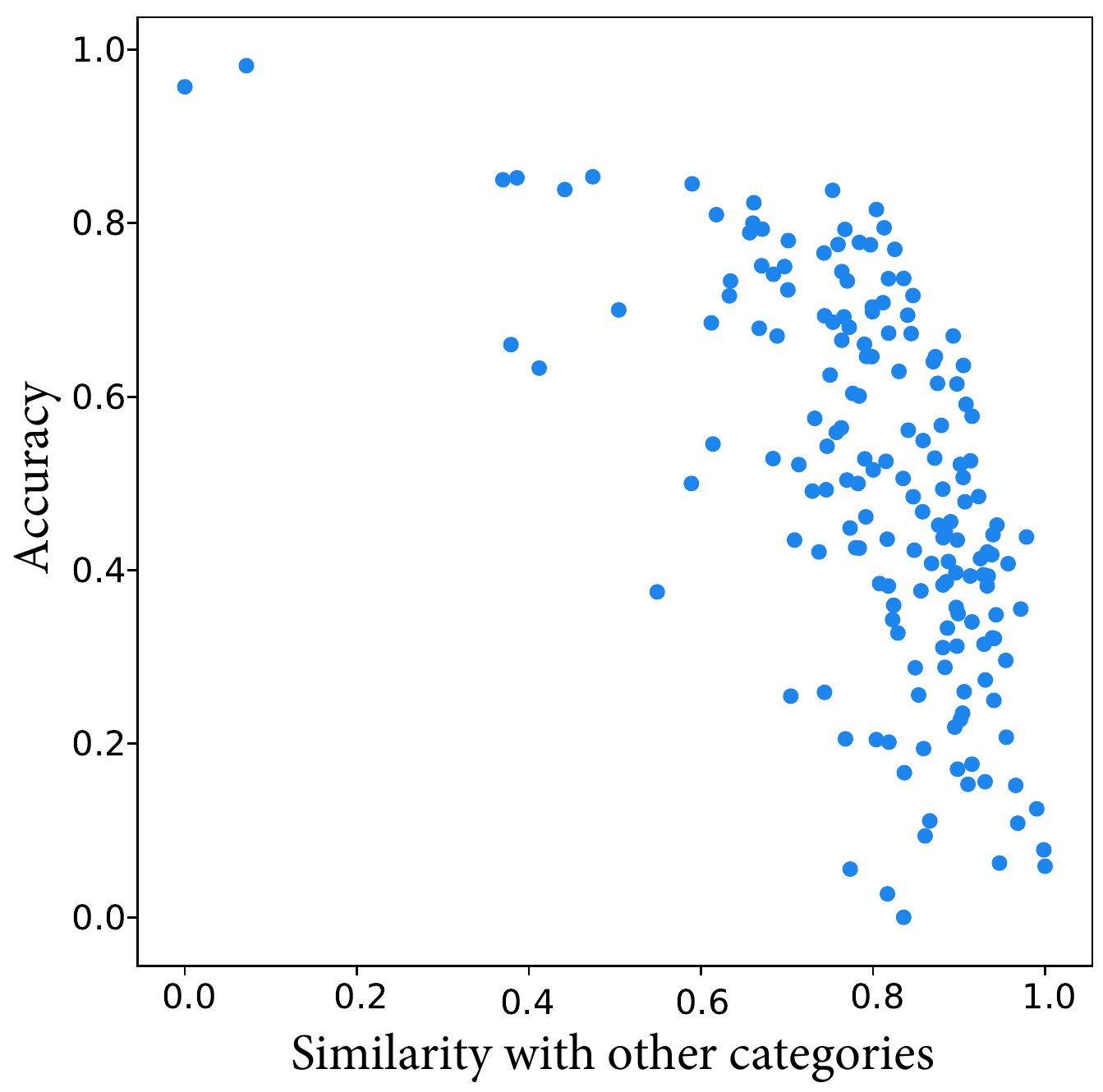}
\end{center}
\caption{Analysis of the influence of feature similarity on category accuracy on the Something-Something dataset. Left: a matrix measuring the feature similarity between different category pairs, in which the color of the grid becomes closer to yellow as the similarity with the corresponding category pair increases. Right: a scatter plot of feature similarity with respect to the category accuracy. {Each point denotes a category, and the abscissa shows the average similarity with other categories.} This figure is best viewed in color.}
\label{fig:CSSM}
\end{figure}

Existing action recognition approaches~\cite{simonyan2014very,carreira2017quo,wang2018appearance,zolfaghari2018eco} usually make one single forward pass to arrive at the final decision. Such approaches, called one-pass methods hereafter, attempt to represent action features from a universal perspective. These one-pass methods fail to consider the relationship between action categories. Intuitively, humans usually deal with the above mix-grained recognition problems by first narrowing down the probable classes and then applying a ``{\em rethinking}'' mechanism towards finer-level recognition. For learning algorithms, the necessity of the ``{\em rethinking}'' mechanism has been verified. Fig.~\ref{fig:CSSM} shows a matrix measuring the category similarity based on the features extracted for action recognition~\cite{xie2018rethinking}. It can be seen that similarities among different categories can vary significantly.
{Moreover, the average interclass feature similarity, which is the average similarity between one class and all the others, is shown to be inversely correlated with the per-class accuracy.} This phenomenon suggests that the universal feature itself is not sufficient to discriminate each class from the classes it is most similar to. Thus, we need class-specific features offering additional discriminability to distinguish similar classes by ``{\em rethinking}''.

\begin{figure*}
\begin{center}
  \includegraphics[width=0.85\linewidth]{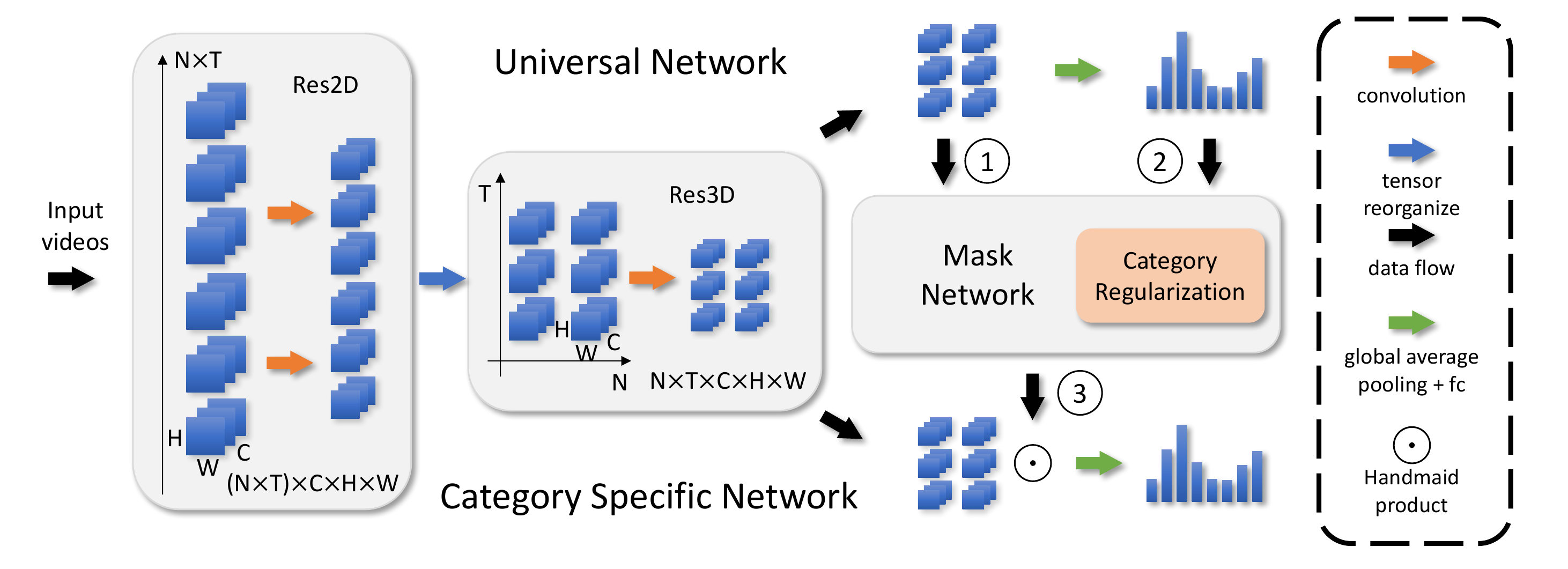}
\end{center}
\caption{The overall framework of the U2S model. The U2S framework consists of three subnetworks: a universal network (UN), a mask network (MN), and a category-specific network (CSN). The UN and CSN have the same architecture and are bridged by the MN. {The UN and CSN share the same Conv2D parameters and have their own Conv3D parameters.}
The UN employs a basic Res2D+Res3D architecture.
The MN integrates feature tensors and predictions of the UN to learn a set of category-specific feature masks using category regularization. These masks are intended to filter out the discriminative features among actions. Guided by the MN, the CSN focuses on the more fine-grained features among confusing categories. This figure is best viewed in color.}
\label{fig:model}
\end{figure*}

In this paper, we present the universal-to-specific (U2S) framework, namely, an end-to-end framework for discriminative feature learning, which consists of three subnetworks, {\em i.e.}, a universal network (UN), a mask network (MN) and a category-specific network (CSN). The overall framework is shown in Fig.~\ref{fig:model}.
The three subnetworks cooperate following the principle of using universal features to guide category-specific features. The universal network is designed to focus on universal signals to separate general conceptions; the mask network 
is regularized with the interclass relationship, approximately defined by the learned universal features and network parameters, to generate a set of category-specific feature masks; and the category-specific network takes the category-specific feature masks from the mask network and focuses more on features corresponding to subtle category-specific differences, thus enabling the separation of fine-grained categories. During prediction, the outputs from both the universal and the class-specific networks are combined into the final prediction.

We perform experiments on three widely used benchmark datasets: the Something-Something, UCF101, 
and HMDB51 datasets. The U2S model consistently outperforms the one-pass models, with top-1 classification accuracies (RGB modality) of 1.95\%, 3.15\% and 2.94\% on the Something-Something, UCF101, and HMDB51 datasets, respectively. Moreover, the universal network and category-specific network are trained collaboratively, leading to an increased discriminative capability of the universal network as manifested by its increased accuracy in an ablation study. In addition, visualizing the feature-level similarity with respect to the class-level accuracy reveals that our approach indeed improves the discriminability of learned features.
The main contributions of this paper are summarized as follows.
\begin{itemize}

\item We propose a novel end-to-end, trainable U2S framework that embeds category-specific feature learning into general feature representations.

\item Our experimental results show that the proposed U2S model consistently improves the feature discriminability and achieves state-of-the-art performance.

\item We introduce a visualization method to compare the feature discriminability and reveal the relationship between the accuracy and the feature discriminability.
\end{itemize}

\section{Related Work}

Action recognition \cite{wang2016temporal,carreira2017quo,li2018videolstm} and action localization \cite{zeng2019graph,zeng2019breaking,chen2019relation} in video have been hot topics in computer vision. Recent methods based on deep learning can be summarized in the following three directions: 3D methods, two-stream methods, and RNN methods.

\subsection{3D models}
The first line of research was inspired by the great success of CNN models~\cite{lecun1998gradient,simonyan2014very,he2016deep} in image recognition tasks and directly extended the convolution kernels, including the temporal dimension. Ji \emph{et al.} \cite{ji20133d} was the first to attempt to handle both the spatial and the temporal dimensions by performing 3D convolution. Tran \emph{et al.} \cite{tran2015learning} designed a deeper network structure (C3D), which achieved better results on the classic datasets UCF101~\cite{soomro2012ucf101} and HMDB51~\cite{kuehne2011hmdb}. Limited by the amount of video data and the large number of parameters introduced by 3D convolution, it is very difficult to train a 3D model from scratch. Then, Carreira \emph{et al.} \cite{carreira2017quo} proposed an I3D model that expands the filters and pooling kernels to 3D with parameters pretrained on ImageNet~\cite{deng2009imagenet}. By leveraging an ImageNet warm start, I3D achieved great performance on the Kinetics dataset~\cite{kay2017kinetics}, and the model pretrained on the Kinetics dataset could further improve the performance on the UCF101 and HMDB51 datasets. Although we can train 3D models such as I3D, these networks have a very large numbers of parameters. It is often difficult to train these models from scratch. To overcome the above bottlenecks, P3D~\cite{qiu2017learning} decomposes the 3D convolution kernels into 2D and 1D convolutions, which reduces the parameters to a large extent. 
Furthermore, (2+1)D~\cite{tran2018closer} uses different combinations of convolutions: spatial followed by temporal, spatial in parallel with temporal, and spatial followed by temporal with a skip connection. Moreover, Xie \emph{et al.} \cite{xie2017rethinking,xie2018rethinking} explored the necessity of using 3D convolutions in all the layers of different depths and proposed using 3D convolutions in selected layers. They also compared the speed and performance between ``{\em top-heavy}'' and ``{\em bottom-heavy}'' networks. Similar to the ``{\em top-heavy}'' idea, ECO \cite{zolfaghari2018eco} is a lightweight convolution network that applies 2D convolutions on the bottom feature maps and 3D convolutions on the top feature maps. This structure can speed up the inference time and even handle online prediction tasks. In addition to these basic convolution models, some novel operation modules were also explored with this model. Wang \emph{et al.} \cite{wang2018appearance} introduced high-order information into their convolution blocks. Wang \emph{et al.} \cite{wang2018non} proposed a nonlocal network to capture long-range dependencies. Zhao \emph{et al.} \cite{zhao2018trajectory} designed a trajectory convolution to integrate features with deformable convolution kernels.

\subsection{Two-Stream models}
The second branch of research concentrates on how to integrate the information of various patterns. Simonyan \emph{et al.}~\cite{simonyan2014two} proposed a two-stream ConvNet architecture that processes RGB frames and the stacked optical flow individually and fuses the classification probability at the end. This was the first attempt to demonstrate that merging the complementary information on appearance and motion can greatly improve performance. Therefore, many works~\cite{wu2015modeling,wang2016temporal,wang2017two,carreira2017quo,huang2018toward,gan2018geometry} have followed the ``two-stream'' idea. Wu \emph{et al.}~\cite{wu2015modeling} combined the two-stream framework and LSTM with a fusion network for video classification. Other frameworks, such as TSN~\cite{wang2016temporal} and I3D~\cite{carreira2017quo}, equipped with a two-stream structure, can achieve great improvements and state-of-the-art performance on action recognition tasks. However, the extraction of optical flow consumes  a massive amount of computing and storage resources. Some researchers have explored alternatives to optical flow. Sevilla \emph{et al.}~\cite{sevilla2017integration} noted that the contribution of optical flow is invariant only to appearance. Therefore, recent studies have preferred to find alternatives to optical flow, such as motion features in \cite{fan2018end}, motion filters in \cite{lee2018motion}, motion priors in \cite{huang2018makes} and motion vectors in \cite{wu2018compressed, zhang2016real, zhang2018real, jiang2019stm}, which can also provide motion information similar to optical flow. In addition to finding alternatives to optical flow, \cite{zhu2018hidden} embedded the computation of optical flow into a convolutional neural  network, which is called a hidden two-stream CNN. Thus, the appearance and motion information can be trained in an end-to-end model.

\subsection{RNN models}
The third active research direction is to use a recurrent neural network (RNN) or its variants, such as long short-term memory (LSTM) and gated recurrent units (GRUs), to handle sequence video data \cite{ng2015beyond,gan2016you,shi2017sequential,zhu2018continuous,agethen2019deep,zhang2018fusing}. The recurrent structure is quite appropriate for dealing with sequence data such as video frames. Ng \emph{et al.} \cite{ng2015beyond} and Shi \emph{et al.} \cite{shi2017sequential} used LSTM cells to deal with frame features that are the output of 2D convolutional neural networks. Sun \emph{et al.} \cite{sun2017lattice} proposed an L2STM model, which extended LSTM by learning independent hidden state transitions of memory cells for individual spatial locations. 
To adapt the basic LSTM structure to handle spatiotemporal data, \cite{xingjian2015convolutional} and \cite{li2018videolstm} changed the state vector to a feature map and changed the fully connected operation to a convolution operation in the LSTM structure.



\subsection{Discussion and Relationship to Prior Work}

{Classification tasks, especially fine-grained settings, usually focus on universal and specific (global and local, respectively) feature representation. The methods in  \cite{huang2016part,zhang2016spda,wei2016mask} localize discriminative parts based on part detection that can utilize the specific features for recognition tasks. However, these local parts are usually predefined, which may require additional annotations. Other works \cite{fu2017look,zheng2019looking,long2018attention} aimed at localizing discriminative parts are based on an attention mechanism that focuses on iterative learning, interchannel relationships, or local feature integration. Moreover, \cite{gan2015devnet,wang2016actions,wang2018videos} explored discriminative parts through spatial-temporal relationships. Existing methods of localizing discriminative parts focus less on the interaction between category relationships in recognition tasks than the framework proposed herein.}

{The closest work to ours is the recently proposed spatial-temporal discriminative filter banks of Martínez \emph{et al}. \cite{martinez2019action}, who introduced the local features by training a set of local discriminative classifiers. In their case, they also combined the global and local features in fine-grained recognition tasks but focused less than our framework on the interaction between category relationships. Moreover, feedback networks from previous works~\cite{zamir2017feedback} are formed in an iterative manner based on feedback received from the previous iteration's output. These networks simply feed the output as a part of the input to an LSTM model.}

{In contrast to these prior methods, we use not only the output but also the category relationships to propose the U2S framework, which rethinks the classification result from universal features and uses category similarities to generate specific features to improve feature discrimination.}


\section{Approach}


In a classification task, what factor can be used to distinguish different categories? For example, when comparing a ``{\em dog}'' and a ``{\em bird}'', the shapes are different, but when comparing a``{\em dog}'' and a ``{\em wolf}'', the tail is one discriminating factor. This factor varies when recognizing different categories of ``{\em dogs}''.
{Categories in a classification dataset are not always equally divided. Some are easy to classify, while some are hard to distinguish since they are quite similar. Therefore, we use the word ``universal'' to denote features that can separate a general concept and the word ``specific'' to denote features containing subtle differences. Existing action recognition baselines~\cite{simonyan2014very,carreira2017quo,wang2018appearance,zolfaghari2018eco} often make one single forward inference to reach the final decision and treat each category equally. Such a one-pass classification approach attempts to learn a discriminative feature representation in a universal feature space. However, this kind of feature representation often shows limited ability to reflect subtle differences among categories. It is thus desired to exploit interclass similarities among categories and learn not only universal features but also category-specific features.}

Previous works aimed at specific or local features~\cite{huang2016part,zhang2016spda,wei2016mask,zhang2017discriminative,liu2017robust,li2018unified} were often based on part detection, a sampling strategy, or some attention mechanism. These methods focused less on the category information and the interaction between universal and specific feature representation than our framework. When facing indeterminate choices, humans may first select some candidates and rethink them from a specific viewpoint. Inspired by the above human decision-making process, we design a novel universal-to-specific (U2S) framework, which uses universal features to guide category-specific features.


\subsection{The U2S Framework}

The overall framework is shown in Fig.~\ref{fig:model}.
The model consists of three subnetworks, namely, a universal network (UN), a category-specific network (CSN), and a mask network (MN); the former two subnetworks share the same architecture and are bridged by the third.
The universal network provides general feature representation and makes the first classification decision. Then, the general features and classification logits are fed into the mask network to help generate feature masks. Finally, features from the category-specific network and masks from the mask network are combined to make the second classification.
The key component of this learning framework is the mask network, which learns a set of category-specific feature masks via a proposed novel method called category regularization.
These masks aim at selecting discriminative features from a more specific viewpoint via the category similarities. With the help of the mask network, the category-specific network can focus on features at a finer level, which contributes to distinguishing between confusing categories ({\em e.g.},``{\em bending something so that it deforms}'' and ``{\em bending something until it breaks}'').

\vspace{4pt}
\noindent
$\bullet$\quad\textbf{Subnetworks: UN and CSN}
\vspace{4pt}

The UN and CSN, shown in Fig.~\ref{fig:model}, are two subnetworks of the U2S framework. The inputs of both the UN and the CSN are videos ({\em e.g.}, RGB frames or stacked optical flow), and their outputs are the classification logits.
Recent action recognition frameworks, especially 3D networks, have achieved great success. The methods in \cite{xie2017rethinking} and \cite{xie2018rethinking} involve several 3D structures and make a trade-off between accuracy and efficiency. Thus, we build our network based on that in~\cite{xie2018rethinking} with its ``{\em top-heavy}'' idea. This design structure only applies 3D convolutions to the top layers and operates smaller feature maps than low-level feature maps due to spatial pooling. Compared with all 3D convolution structures, this 2D plus 3D convolution can reduce the computational cost. Thus, our architecture has a  ResNet~\cite{he2016deep} backbone with the original Res2D and its variation in the Res3D structures. Following the basic structure notations in~\cite{he2016deep}, the main convolution components are residue groups from group p0 to group 3. In our universal network and category-specific network, the transition position from 2D to 3D is set between group 1 and group 2. Stacking the temporal dimension, the output tensor $(NT\times C\times  H\times W)$ from the Res2D network is reorganized and transposed to five dimensions $(N\times T\times C \times H\times W)$. $N$, $T$, $C$, $H$, and $W$ represent the batch size, temporal length, feature channel, image height, and image width, respectively. The final output logits from the UN and CSN are used to compute the cross-entropy loss (notated as $\mathcal{L}_{\mathrm{U}}$ and $\mathcal{L}_{\mathrm{C}}$).
In addition, preactivated batch normalization~\cite{he2016identity} is applied to all the convolution layers.

\vspace{4pt}
\noindent
$\bullet$\quad\textbf{A Mask Network as the Bridge}
\vspace{4pt}

\begin{figure}[t]
\begin{center}
  	\includegraphics[width=0.75\linewidth]{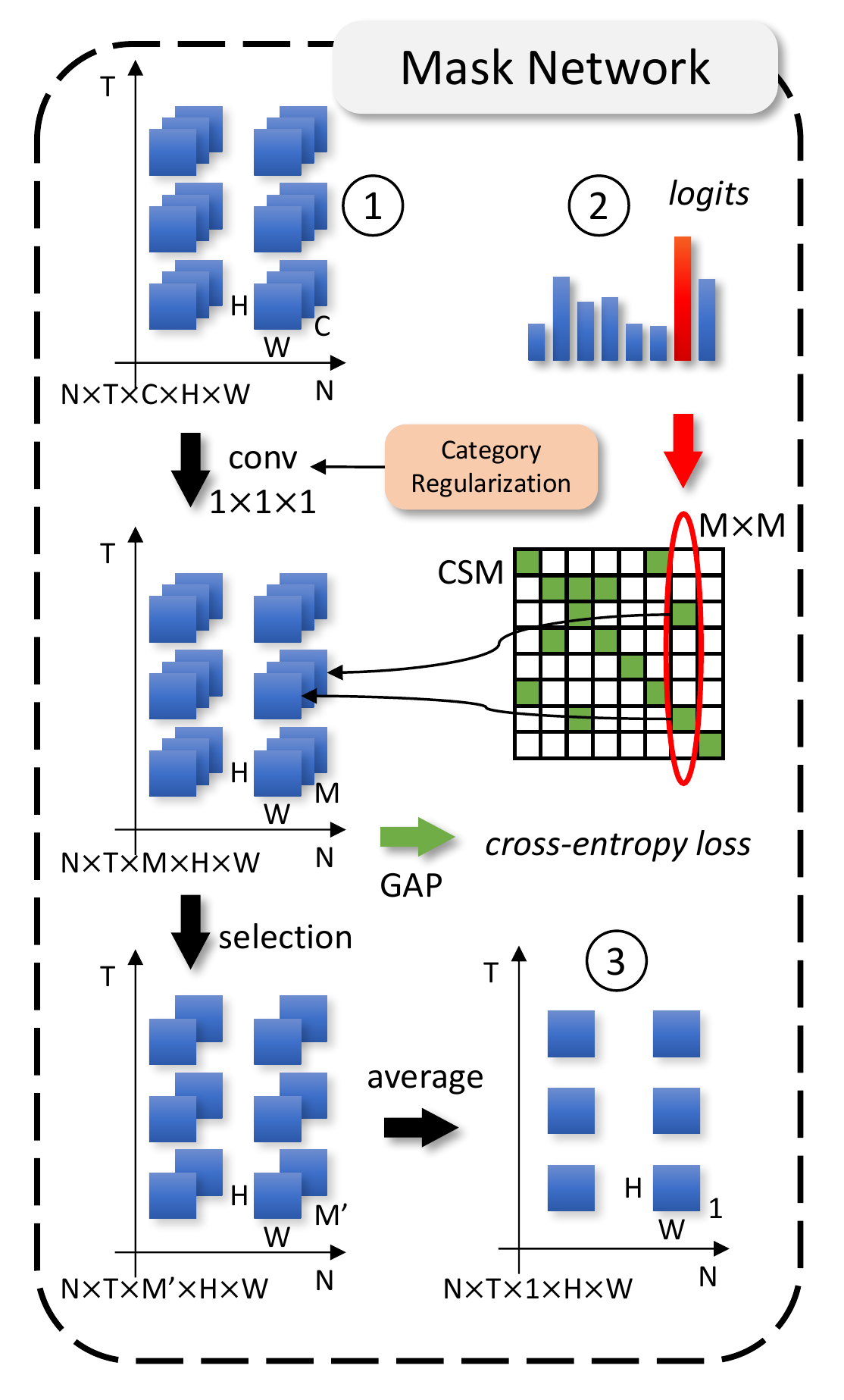}
\end{center}
\caption{The proposed mask network. The MN targets learning a set of category-specific feature masks that can select confusing category features according to the prediction of the universal network. \ding{192} is the input tensor $(N\times T\times C \times H\times W)$ from the universal network. \ding{193} are the predicted logits of the universal network, which can help to select the category subset of the confusing categories. \ding{194} is the combination of selected feature masks. The green blocks in the CSM (binary) represent positive values. This figure is best viewed in color.}
   \label{fig:MN}
\vspace{-4pt}
\end{figure}

The mask network learns the feature masks to help the CSN focus on the features at a finer level. The mask network first provides masks for each category; then, according to the prediction from the UN, the masks of confusing categories are selected and combined as the output mask.
Fig.~\ref{fig:MN} shows the detailed pipeline for the mask network. The numbers enclosed in circles correspond to the inputs and outputs of the mask network shown in Fig.~\ref{fig:model}. Input circles 1 and 2 are the feature map tensor $(N\times T\times C\times H\times W)$ and the predicted logits from the universal network, respectively. Output circle 3 represents the final mask. For a detailed explanation of the mask network and how it can generate category-specific masks, we first introduce the category similarity matrix (CSM) and then describe the mask generation pipeline.

\noindent \textbf{CSM}. We denote the CSM as $\mathbf{C}\in\{0,1\}^{M\times M}$, where $M$ is the number of categories. A positive value in the CSM means that the two corresponding categories are easily confused.
First, we introduce the method to measure the category feature similarity. Inspired by the visualization method in~\cite{kalantidis2016cross}, the ``\textit{sparsity pattern of channels contains discriminative information}''. To utilize this sparsity pattern, we first extract convolutional feature maps. Here, we extract the output of group 3 from the UN $(N\times T\times C\times H\times W)$. In the following equations, we use superscript $c$ to denote the channel and the subscript $m$ to denote the category.
For each channel $c$ in the feature maps, we spatiotemporally count the proportion of nonzero responses $Q^{c}$ and calculate the per-channel sparsity $\Xi^{c}$ as follows:
\begin{equation}\label{eq:channel-sparsity}
   \Xi^{c} = 1-Q^{c}.
   \end{equation}
All the channel sparsities $\Xi^{c}$ constitutes a vector $\mathbf{\Xi}\in[0,1]^{C}$ to represent the feature response. For different categories, we calculate the average feature sparsity $\mathbf{\Xi}$ of all the samples under each category to represent this category-specific feature response $\mathbf{\Xi}_{m}$.
Finally, utilizing the distance function $f_{\mathrm{dist}}: R^C \times R^C \rightarrow \{x | x \in [0, 1], x \in R\}$, we calculate similarities between different categories with $\{\mathbf{\Xi}_{m}\},m=1,2,...,M$, where $M$ is the number of categories:
\begin{equation}\label{eq:similarity}
   s_{i,j} = 1-f_{\mathrm{dist}}(\mathbf{\Xi}_{i},\mathbf{\Xi}_{j}),
   \end{equation}
where $s_{i,j}$ stands for similarity between category $i$ and category $j$. Thus, we obtain a matrix $\mathbf{S}\in[0,1]^{M\times M}=[s_{i,j}]$ to measure the category similarities. To simplify the screening process for highlighting confusing categories, we denote the CSM ($\mathbf{C}\in\{0,1\}^{M\times M}$) as a binary version of $\mathbf{S}\in[0,1]^{M\times M}$ by a constant threshold $\alpha$. Then, the CSM can be used to represent the category similarities.

\noindent \textbf{Mask Generation}. As shown in Fig.~\ref{fig:MN}, to learn a set of category-specific masks, a convolution operation with a $conv 1\times 1\times 1$ kernel is applied, which can be regarded as a linear feature combiner. The number of output channels is set to be $M$ (the number of categories). Therefore, each feature map in the channel dimension can be regarded as a category-specific mask.
To ensure that these learned category-specific masks $(N\times T\times M\times H\times W)$ are activated according to their categories, we use classification labels to train them.
In detail, these masks are processed by global average pooling (green array); then, the mask tensors are converted to logit form $(N\times M)$ and are then used to compute the cross-entropy loss (denoted by $\mathcal{L}_{\mathrm{M}}$). To further constrain the discriminability between masks, the category regularization, applied to parameters of the $conv 1\times 1\times 1$ layer, is introduced and will be explained in the next subsection.

Once we obtain the category-specific masks, how do we obtain the final mask? For the right panel of Fig.~\ref{fig:MN}, when the predicted category $i$ of the universal network (circle 2) is fed into the mask network, the confusing categories are selected via positive values in the \emph{i-th} column of the CSM. Therefore, we select these corresponding masks from
among $(N\times T\times M\times H\times W)$ masks in three dimensions. 
Finally, each video in batch $N$ obtains a combined (averaged) feature mask $(N\times T\times 1\times H\times W)$ as the final output mask. If the prediction (circle 2) does not have confusing categories, the mask network is degraded into a self-attention module as in~\cite{miech2017learnable}, which can also help the category-specific network select valuable features.

\subsection{Loss Function}

For the action recognition task, we use the cross-entropy loss to train our network, and the loss function is defined as:
\begin{equation}\label{eq:loss}
   \mathcal{L} = \mathcal{L}_{\mathrm{U}}+\mathcal{L}_{\mathrm{C}}+\mathcal{L}_{\mathrm{M}}+\lambda w_{\mathrm{regular}},
\end{equation}
where $\mathcal{L}_{\mathrm{U}}$, $\mathcal{L}_{\mathrm{C}}$, and $\mathcal{L}_{\mathrm{M}}$ are the cross-entropy losses from the universal network, category-specific network, and mask network, respectively. $w_{\mathrm{regular}}$ denotes category regularization, and $\lambda$ is added to control the regularization weight.

Category regularization is used to constrain the discriminability between category-specific masks. These masks are learned from the input features of the mask network with the parameters of the $conv 1\times 1\times 1$ layer, which is a matrix $\mathbf{W}\in\mathbb{R}^{C\times M}$. $C$ is the input feature dimension, and $M$ is the number of categories. The $conv 1\times 1\times 1$ layer can be regarded as a linear feature combiner.
Therefore, a vector ($\mathbf{w}_{i}$) of the {\em i-th} column in $\mathbf{W}$ decides how to generate the {\em i-th} category-specific mask.
Therefore, we can constrain the similarity between $\{\mathbf{w}_{i}\}$ to vary the category-specific mask.

As shown in Eq.~\ref{eq:regularization1}:

\begin{equation}\label{eq:regularization1}
  \mathbf{S}^{w}= 
  \left[                
     \begin{array}{ccc}   
        s(\mathbf{w}_{1},\mathbf{w}_{1})  
      & s(\mathbf{w}_{1},\mathbf{w}_{2}) 
      & ...\\ 
        s(\mathbf{w}_{2},\mathbf{w}_{1}) 
      & s(\mathbf{w}_{2},\mathbf{w}_{2}) 
      & ...\\ 
        ... 
      & ... 
      & s(\mathbf{w}_{M},\mathbf{w}_{M})\\  
     \end{array}
  \right],      
\end{equation}

\noindent where $\mathbf{S}^{w}\in[0,1]^{M\times M}=[s^{w}_{i,j}]$ is the weight similarity matrix. Similar to Eq.~\ref{eq:similarity}, $s(\mathbf{w}_{i},\mathbf{w}_{j})$ denotes the similarity between $\mathbf{w}_{i}$ and $\mathbf{w}_{j}$. Then, we can obtain $w_{\mathrm{regular}}$:

\begin{equation}\label{eq:regularization2}
  w_{\mathrm{regular}} = \frac{1}{N_{\mathrm{pos}}} \sum_{i,j} s^{w}_{i,j}.
\end{equation}

\noindent In Eq.~\ref{eq:regularization2}, we only constrain the weight similarity between the confusing categories by the CSM ($\mathbf{C}$). ${N_{\mathrm{pos}}}$ is a normalized constant denoting the number of positive values excluding the diagonals of the CSM. Thus, the selected pairs can be represented by positive positions in $\mathbf{M}\in\{0,1\}^{M\times M}=[m_{i,j}]$:
\begin{equation}\label{eq:regularization3}
  \mathbf{M} = \mathbf{C} - \mathbf{I}.
\end{equation}

\noindent Since $s(\mathbf{w}_{i},\mathbf{w}_{j})$ depicts the weight similarity between the \emph{i-th} and \emph{j-th} categories, we constrain the weight similarity via filtering out the confusing category pairs using $\mathbf{M}$, $(i,j)\in\{(i,j)|m_{i,j}=1\}$.





\section{Experiment}

\subsection{Datasets}
We mainly conduct our experiments on the Something-Something V2 dataset~\cite{goyal2017something} (called Something-Something hereafter), which is a large collection of densely labeled video clips that show humans performing basic actions with everyday objects. This dataset consists of 220,847 videos with 174 classes. Some of the classes are quite fine-grained and easily confused with each other, such as ``{\em bending something so that it deforms}'' and ``{\em bending something until it breaks}''. Thus, the criteria for classification are mostly determined by some details. In addition, we also conduct our experiments on the UCF101~\cite{soomro2012ucf101} and HMDB51~\cite{kuehne2011hmdb} datasets, which are relatively small action recognition datasets that also contain rich categories. The UCF101 dataset is composed of 13,320 realistic user-uploaded video clips and 101 action classes such as ``HorseRiding'', ``Skiing'', ``Drumming'' and ``Archery''. The definition of categories is mainly focused on sports, music, and some daily behaviors. The average clip length is approximately 7 seconds with a frame rate of 25 fps. The HMDB51 dataset is from a variety of sources ranging from digitized movies to YouTube. It contains 6,766 video clips defined by 51 action categories,  such as ``Jump'', ``Kiss'', and ``Laugh''. The definitions of the categories are biased towards some simple daily actions. {Moreover, we also implement our proposed U2S framework on two fine-grained image classification datasets, Stanford Cars~\cite{KrauseStarkDengFei-Fei_3DRR2013} and Stanford Dogs~\cite{KhoslaYaoJayadevaprakashFeiFei_FGVC2011}. The Stanford Cars dataset contains 16,185 images of 196 classes of cars, which are divided into 8,144 images for training and 8,041 images for testing. The Stanford Dogs dataset contains 20,580 images of 120 classes of dogs from around the world, which are divided into 12,000 images for training and 8,580 images for testing.}

\subsection{Implementation Details}

\noindent \textbf{Frame Sampling}. As illustrated in Fig.~\ref{fig:model}, video clips with a variable number of frames are fed into the network. To address the different frame lengths, the video is split into N equal subsections, and the frames are sampled at the same position in each subsection. This strategy maintains the movement rate between adjacent frames. In the training phase, positions in a subsection are randomly selected, enabling the network to fully exploit all the frames and provide more diversity during training. When testing the action videos, the position in the subsection is fixed.

\noindent \textbf{Training}. The first training stage is to train the universal network using mini-batch SGD with momentum and utilize preactivated batch normalization in each convolution layer. Assisted by this well-trained universal network, we obtain the CSM based on the corresponding dataset that will be applied to the mask network and category regularization.
The second training stage is to train the category-specific network with masks provided by the mask network, which focuses on discriminative features among easily confused categories.
Finally, we jointly train the whole network with a relatively low learning rate. We sample 18 frames from the input videos following the strategy described in the previous section. In addition, the short edges of the input frames are resized to 240 and randomly cropped to $224\times 224$ at the same location for the frames in a whole video clip. However, the frequently used flipping augmentation strategy is not applied due to some mirror-sensitive categories defined in the Something-Something dataset, such as descriptions with specific direction words (``left and right'', ``up and down'').
{The threshold $\alpha$ for the CSM determines how many categories a category can be confused with and thus depends on the dataset itself. The straightforward way to determine this constant threshold is to consult the top-N classification accuracy. We choose the threshold by letting the average number of confusing categories be N if this top-N accuracy can achieve a high standard (85\% for Sth-Sth-V2, 95\% for UCF101 and Stanford Cars, and 90\% for HMDB51 and Stanford Dogs), which means that choosing N categories will include most of the confusing categories.}
In our experiments, the ResNet-50 structure is selected as our backbone. The training hyperparameters are as follows. The initial learning rate is set to $10^{-3}$ for the first and second training stages and decreases by a factor of 10 when the validation error saturates after 7 and 12 epochs. We train the network with a momentum of 0.9, a weight decay of $10^{-4}$, and a batch size of 20. The Res2D and Res3D weights are initialized with pretrained networks on Kinetics. The learning rate of the final joint learning processing step is $10^{-4}$, and the $\lambda$ for category regularization is set to 0.5, balancing the $w_{\mathrm{regular}}$ in the final loss function.

\noindent \textbf{Testing}. During testing, the universal network and category-specific network propagate jointly, with the coupling position at the final feature level bridged by the mask network. After two predictions, we merge the predicted scores in different networks, utilizing the complementarity of the two predictions. The final score is fused by 10 crops, as in~\cite{wang2017temporal,wang2018appearance}.


\subsection{Results on the Something-Something Dataset}

\subsubsection{Ablation studies on fusing the U2S predictions}

In this section, we first discuss the methods for merging all the predictions in our framework. As illustrated in Fig.~\ref{fig:model}, we obtain three classification predictions, namely, universal prediction, bridge prediction, and specific prediction, from the universal network, mask network, and category-specific network, respectively. The universal prediction and bridge prediction both use universal features from the universal network, while the specific prediction uses masked features from the category-specific network. Table~\ref{table:merge} reports the classification results from the one-pass model and all combinations of the universal, bridge, and specific predictions.

The one-pass universal results and universal results are both from the universal network and differ in whether joint training is used (without and with joint training, respectively). We observe a 1.15\% performance gain from 56.52\% to 57.67\% from joint training with the universal network and category-specific network. The universal and bridge results, which both use universal features, achieve a similar performance, which is higher than that of the specific prediction. However, when the specific prediction is fused with any of the merged results, the performance is greatly improved. This finding means that selected category-specific features effectively complement the universal features. All the merged results are the simple average of the prediction scores. In addition, when comparing the performance gain between the top-1 and top-5 accuracies provided by fusing the specific predictions, we observe that the top-5 accuracy is almost unchanged but that the top-1 accuracy has improved. This finding means that the category-specific features provide discriminative information on the most confusing examples. We further illustrate the differences in feature discriminability in the following subsection.

\subsubsection{Ablation studies on category regularization}

The effectiveness of category regularization is shown in Table~\ref{table:category-regularization}. The top-1 classification accuracy on the Something-Something V2 dataset is improved by 1.28 percentage points with category regularization applied. The visualization of weight similarity with or without category regularization will be shown in a later subsection.

\begin{table}[t]
\begin{center}
\caption{The classification results on Something-Something dataset. The results are from the one-pass model and all the combinations of the universal, bridge, and specific predictions. The jointly trained universal and specific models achieve high performance when merging all the predictions.}
\label{table:merge}
\begin{tabular}{ccccc}
\toprule
\multicolumn{3}{c}{Method}             & Top-1 (\%)       & Top-5 (\%)       \\ \hline
\multicolumn{3}{c}{One-Pass Universal} & 56.52          & 82.95          \\ \hline
Universal    & Bridge    & Specific    & \multicolumn{2}{c}{}            \\ \hline
\checkmark           &           &             & 57.67          & 83.96          \\
             & \checkmark        &             & 57.58          & 83.26          \\
             &           & \checkmark          & 55.95          & 81.50          \\
\checkmark           & \checkmark        &             & 57.87          & 83.90          \\
             & \checkmark        & \checkmark          & 58.33          & 83.99          \\
\checkmark           &           & \checkmark          & 58.17          & 83.57          \\
\checkmark           & \checkmark        & \checkmark          & \textbf{58.47} & \textbf{84.03} \\ 
\bottomrule
\end{tabular}
\end{center}
\end{table}

\begin{table}[t]
\begin{center}
\caption{The classification results on the Something-Something-V2 dataset with and without category regularization.}
\label{table:category-regularization}
\begin{tabular}{ccc}
\toprule
Setting                     & Top-1 (\%) & Top-5 (\%) \\ \hline
w/o category regularization & 57.19     & 83.34      \\
w/ category regularization  & \textbf{58.47}     & \textbf{84.03}      \\ 
\bottomrule
\end{tabular}
\end{center}
\end{table}

\begin{table}[t]
\begin{center}
\caption{The ablation studies of the MN on the Something-Something V2 dataset. }
\label{table:ablation_mn}
\begin{tabular}{lccc}
\toprule
\makebox[3cm][c]{Settings}                 & Para. (G)     & Top-1 (\%)          & Top-5 (\%)        \\ \hline
\makebox[3cm][c]{One-Pass Universal}       & 60.77         & 56.52              & 82.95             \\\hline
a) U2S                                     & 120.43        & \textbf{58.47}     & \textbf{84.03}    \\ 
b) U2S (simple)                            & 120.43        & 57.38              & 83.06             \\
c) U2S (soft CSM)                          & 120.43        & 57.48              & 83.48             \\
d) U2S - MN                                & 120.09        & 56.69              & 83.01             \\
e) U2S - MN + Non-Local \cite{wang2018non} & 120.84        & 58.02              & 83.78             \\ \hline
f) U2S (oracle)                            & 120.43        & 63.54              & 88.90             \\ 
\bottomrule
\end{tabular}
\end{center}
\end{table}

\subsubsection{{Ablation studies on the mask network}}

{Our main contribution is to design a mask network to guide more discriminative category-specific features. We deeply investigate this mask network and perform some variant experiments, the results of which are shown in Table~\ref{table:ablation_mn}. First, we provide clear descriptions of these experimental settings: \textbf{a)}: the proposed U2S model in the ``Approach'' section; \textbf{b)}: we simplify our mask network and select only one mask of the predicted category by using the UN, which means that the MN does not consider the masks of other confusing categories; \textbf{c)}: we change the CSM from the binary version ($\mathbf{C}\in\{0,1\}^{M\times M}$) to the absolute version ($\mathbf{C}\in[0,1]^{M\times M}$); \textbf{d)}: we remove the mask network and report the results by using an ``ensemble'' operation of the UN and CSN; \textbf{e)}: we compare another self-attention method, the non-local method~\cite{wang2018non}, by following the same backbone; \textbf{f)}: we use the ground-truth category as guidance for mask generation (not the output of the UN), which can be regarded as the upper bound of the proposed method.}

{Comparing a) and b), it can be seen that considering the masks of other confusing categories increases the performance gain since these masks contain the discriminative spatial-temporal regions between confusing categories. Comparing a) and c) shows that the absolute CSM cannot achieve the performance that the binary CSM can. We postulate the masks of other non-confusing categories will introduce noise into the final mask, which will affect the final performance. Although each non-confusing category mask has a small weight, aggregating them
results in a non-negligible weight 
due to the large number of categories. Comparing a), d), and the one-pass model results, we find that the performance improvement does not come from introducing more parameters since the performance of d) is not obviously improved over that of the two-branch architecture; instead, most of the performance gain is produced by the discriminative features learned by the CSN with the MN (only one convolution layer).}

\subsubsection{Visualizing Feature Similarity Distributions}

\begin{figure*}[t]
\begin{center}
\subfloat[]{
      \includegraphics[width=0.38\linewidth]{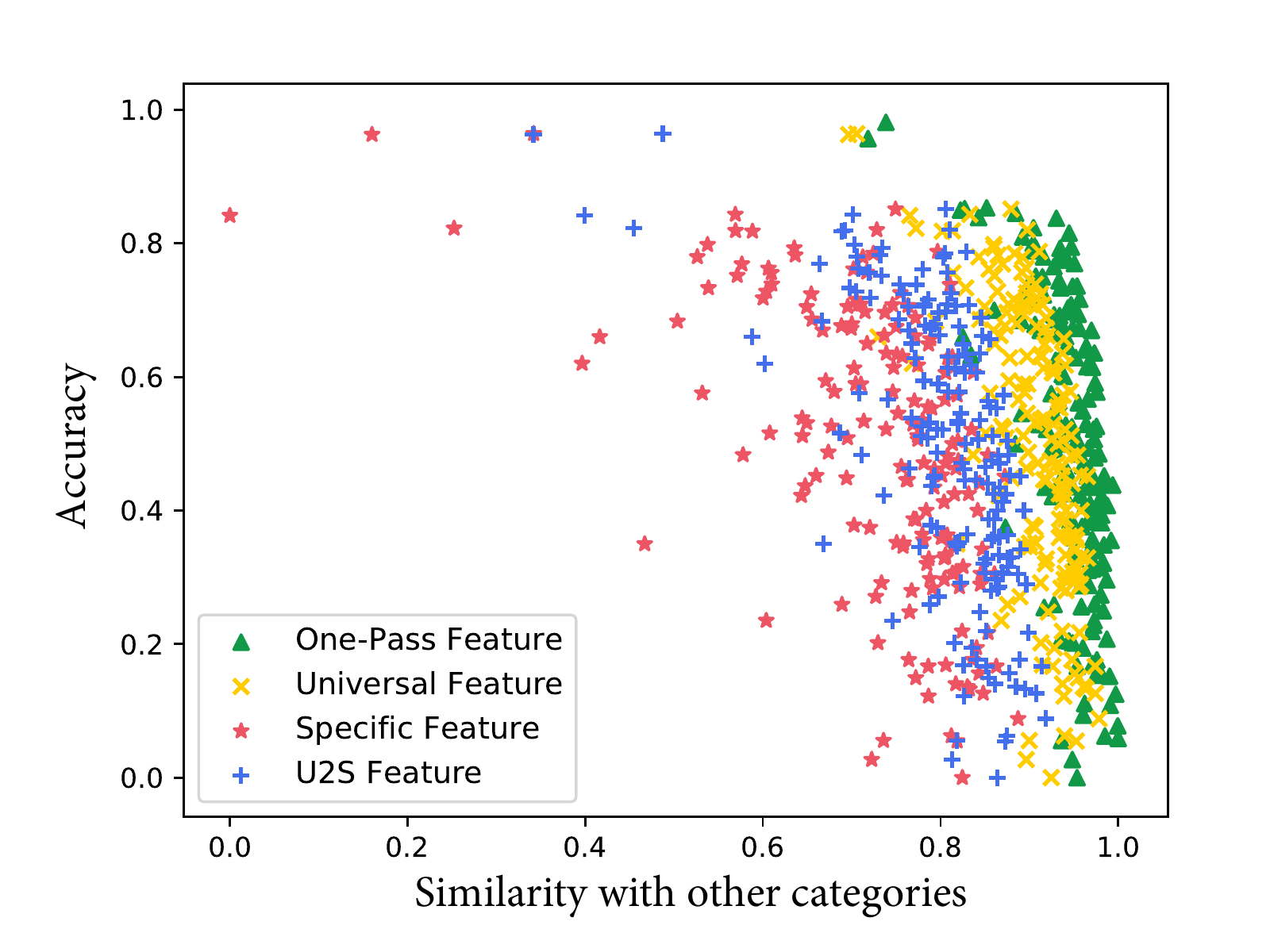}
} \quad
\subfloat[]{
      \includegraphics[width=0.38\linewidth]{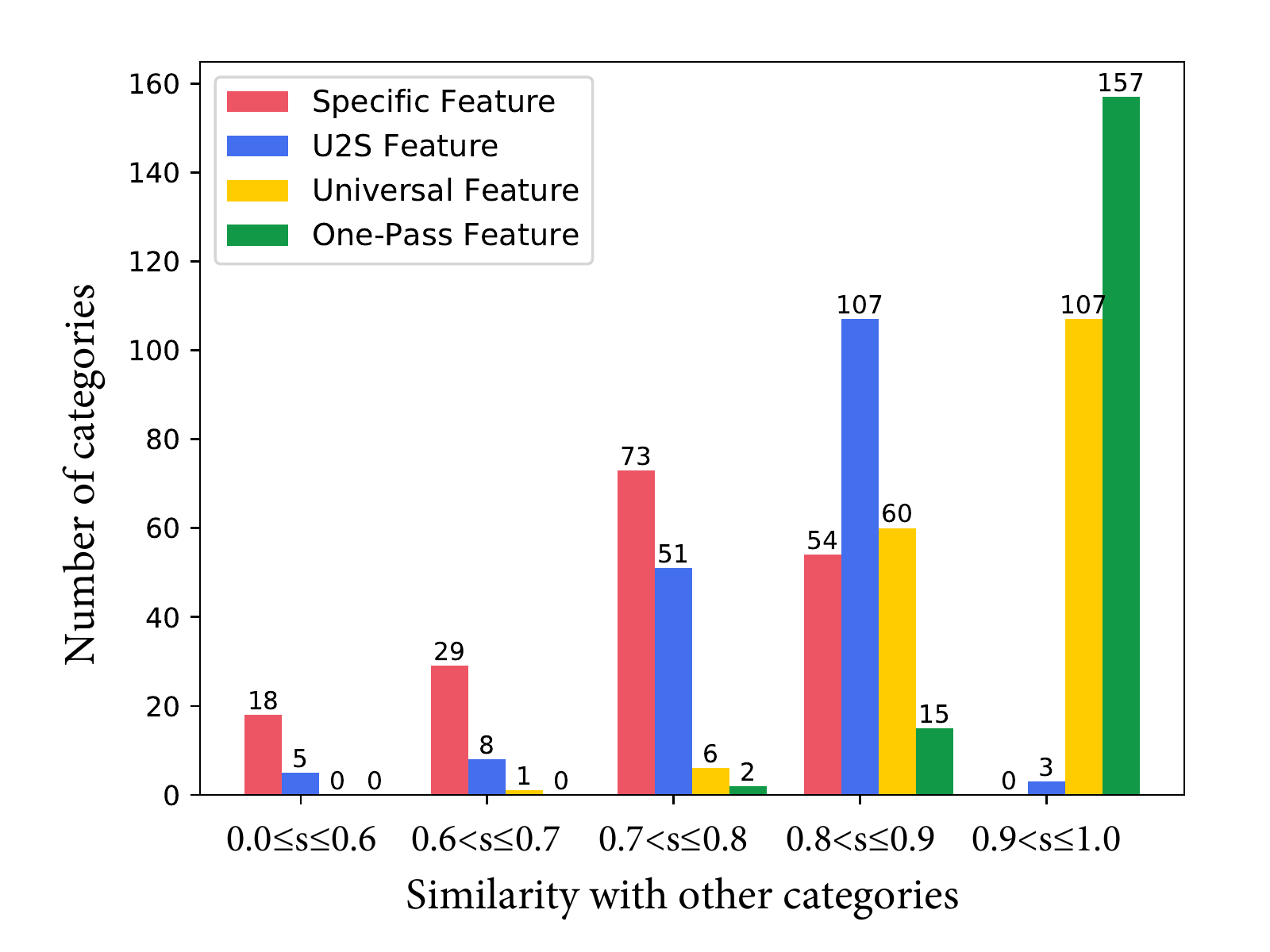}
}
\end{center}

\caption{Feature similarity distributions among the one-pass, universal, specific, and U2S features on the Something-Something dataset. (a) Scatter plot of the feature similarity vs. category accuracy. (b) Histogram of the feature similarity counting the specific number of categories in different similarity intervals. This figure is best viewed in color.}
   \label{fig:feature}
\end{figure*}

The scatter plot illustrated in Fig.~\ref{fig:feature} (a) shows the classification accuracy in each category versus the category similarity with other categories. The similarity with other categories is calculated by $\mathbf{S}\in[0,1]^{M\times M}$ (defined in the description of the CSM in Sec.~\uppercase\expandafter{\romannumeral3}):
\begin{equation}\label{eq:channel sparsity}
   \mathbf{s}_{\mathrm{similarity}}=\mathrm{Norm}(\sum_{j} (\mathbf{S})_{i,j}-\mathbf{I}),
   \end{equation}
where $\mathbf{I}$ is the identity matrix, $\mathrm{Norm}(\cdot)$ is the min-max normalization function, and thus, $\mathbf{s}_{\mathrm{similarity}}\in[0,1]^{M}$. Each value in $\mathbf{s}_{\mathrm{similarity}}$ represents the corresponding averaged interclass similarity with other categories since the self-calculated similarity has been subtracted from the diagonal position.

Fig.~\ref{fig:feature} (a) illustrates the feature similarity distributions among different feature representations. Each point in the scatter plot represents a category. To prove that our category-specific network can learn more discriminative features, we use the min-max normalization method to restrict the feature similarity on a unified scale among four kinds of feature representations. The green triangles, yellow crosses, red stars, and blue plus symbols represent one-pass, universal, specific, and U2S features, respectively. The one-pass features and universal features are both extracted from the UN, but the yellow crosses are distributed on the right-hand side of the plot thanks to the joint learning. Certainly, the red stars are distributed the farthest to the left, which means that the specific features are quite discriminative. However, the use of only specific features cannot achieve a high performance, as shown in Table~\ref{table:merge}, since the selection strategy ignores some global information. Combining the universal features and specific features complements the feature representations in Fig.~\ref{fig:feature} (a) and achieves a high performance, as shown in Table~\ref{table:merge}.
To show the similarity distribution, we plot a histogram to count the number of categories in different similarity intervals. Fig.~\ref{fig:feature} (b) shows that the category-specific network distinguishes more categories than the other networks, and these discriminative features contribute greatly to classification tasks.

\subsubsection{Visualizing Masks}

\begin{figure*}[h!]
\begin{center}
  	\includegraphics[width=0.8\linewidth]{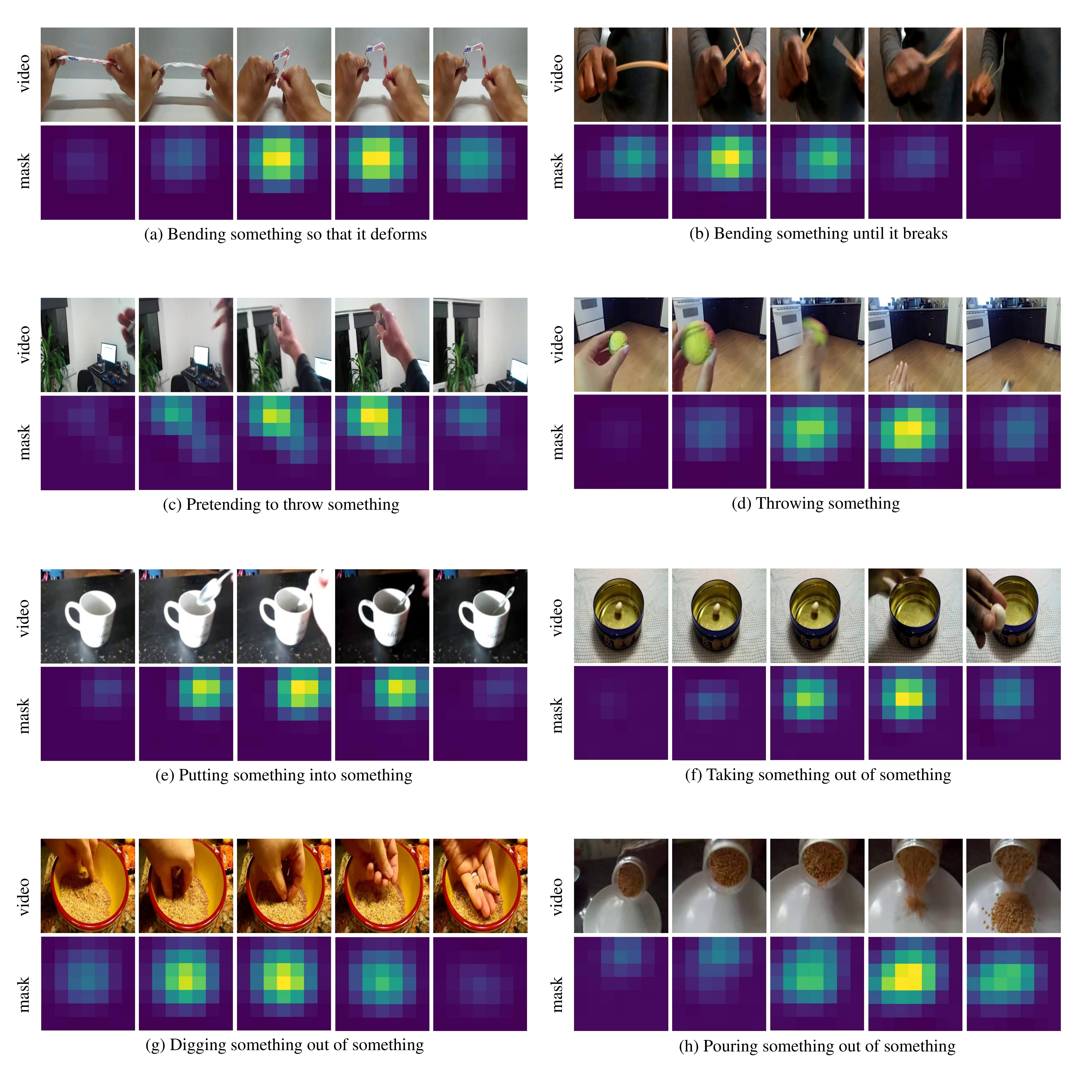}
\end{center}
\caption{The visualization of learned masks from the mask network for similar action pairs -- (a), (b): ``bending something so that it deforms'' and ``bending something until it breaks''; (c), (d): ``pretending to throw something'' and ``throwing something''; (e), (f), (g), (h): ``putting something into something'', ``taking something out of something'', ``digging something out of something'', ``pouring something out of something''.}
  \label{fig:Mask}
\end{figure*}

In this section, we visualize the masks learned from the mask network. During the second stage of our U2S framework, we aim to classify action categories at a relatively fine level. Therefore, finding the most discriminative part between confusing categories is crucial. As illustrated in Fig.~\ref{fig:Mask}, we visualize some videos and their learned masks between confusing categories, which reveals that our approach indeed focuses on the discriminative part.

For example, the videos in Fig.~\ref{fig:Mask} (a) and (b) show actions of ``bending'', namely, ``bending something so that it deforms'' and ``bending something until it breaks''. The key evidence for distinguishing these two actions is whether the bent object has been broken. As shown in the learned masks, our mask network tends to focus on the most important spatial-temporal location of the state of an object when it is bent. Another pair of actions is shown in Fig.~\ref{fig:Mask} (c) and (d), namely, ``pretending to throw something'' and ``throwing something''. These two action categories are easily confused since the difference is whether the object in question is actually thrown. Therefore, we find the highlighted spatial-temporal areas in the masks that focus on the hand after the throwing action. In the remaining similar action categories shown in Fig.~\ref{fig:Mask} (e), (f), (g) and (h), the actions are all about ``something out of or into something''. The learned masks focus on the locations where and when something moves. To further quantitatively demonstrate the performance of the U2S framework, we list the top-1 classification accuracies of these action categories before and after the U2S feature learning process in Table~\ref{table:compare}. The classification accuracy is improved in each confusing action group. It is observed that U2S results in 3.92\% and 4.85\% performance gains on ``bending'' actions; ``10.64\%'' and ``5.44\%'' gains on throwing actions; and an average gain of 8\% on actions involving ``something out of or into something''.

\begin{figure}{}
\centering
\begin{minipage}{1.0\linewidth}
\begin{center}
\includegraphics[width=0.95\linewidth]{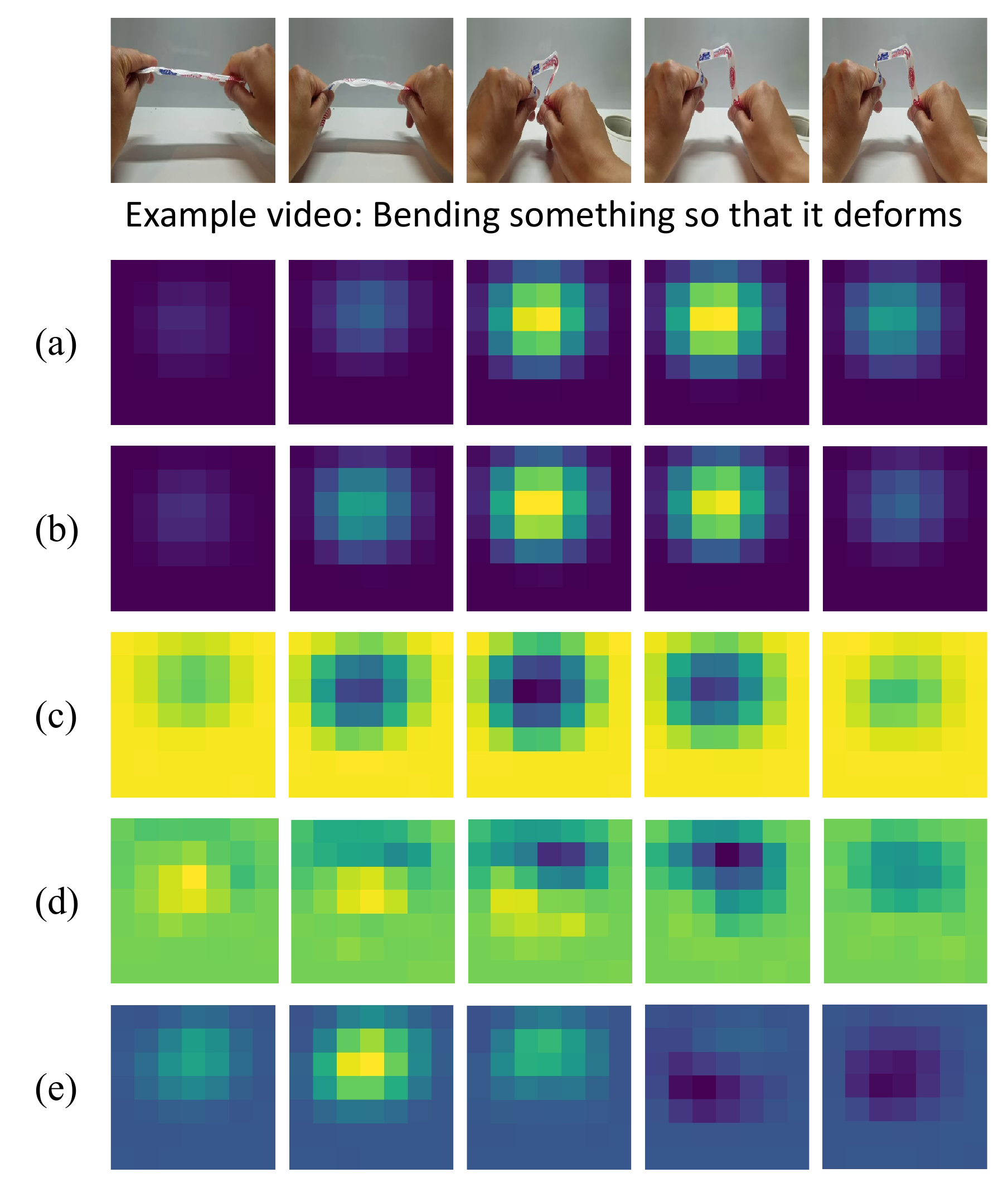}
\end{center}
\end{minipage}
\\
\begin{minipage}{1.0\linewidth}
\quad \quad \quad
\begin{center}
\includegraphics[width=0.44\linewidth]{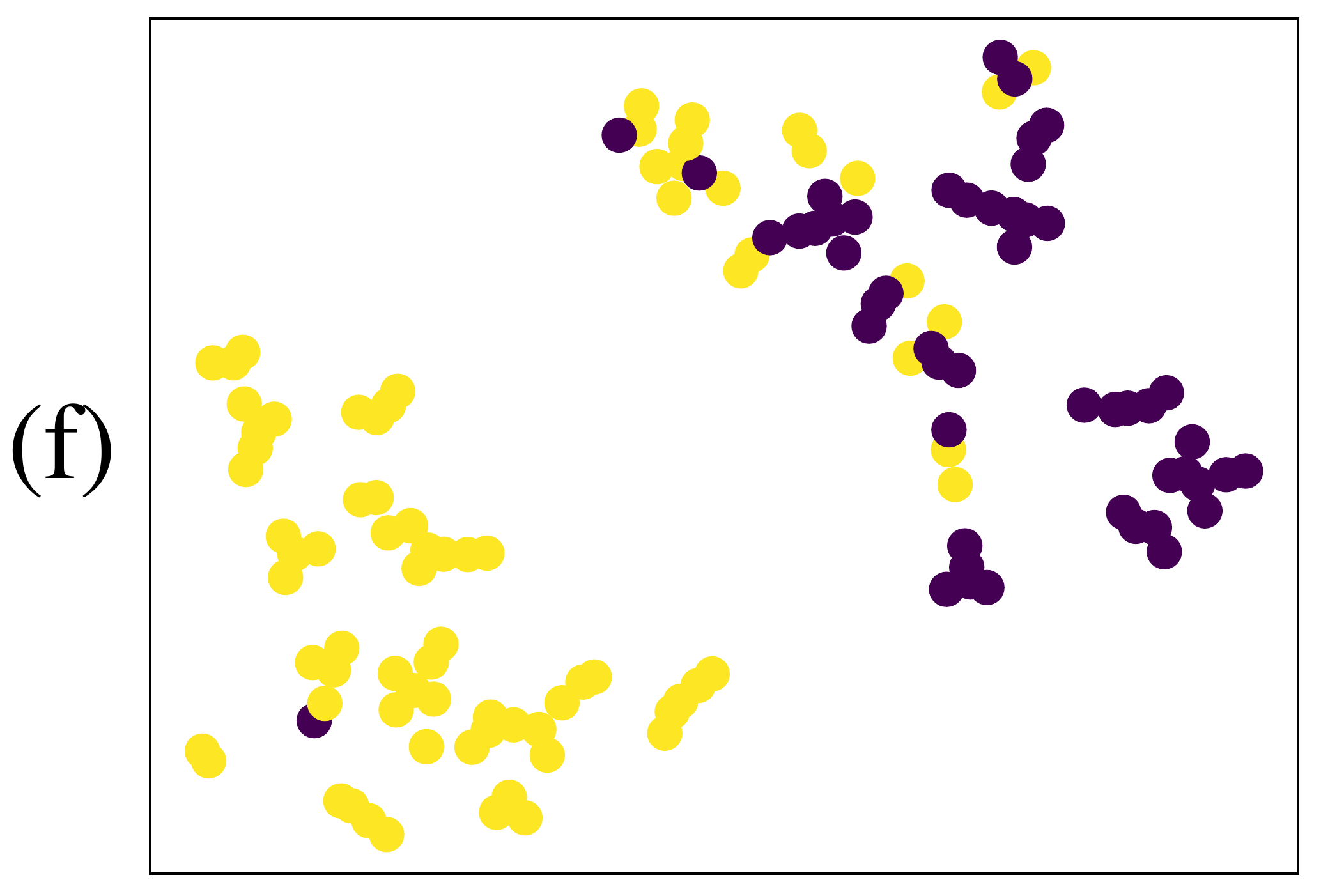}
\includegraphics[width=0.44\linewidth]{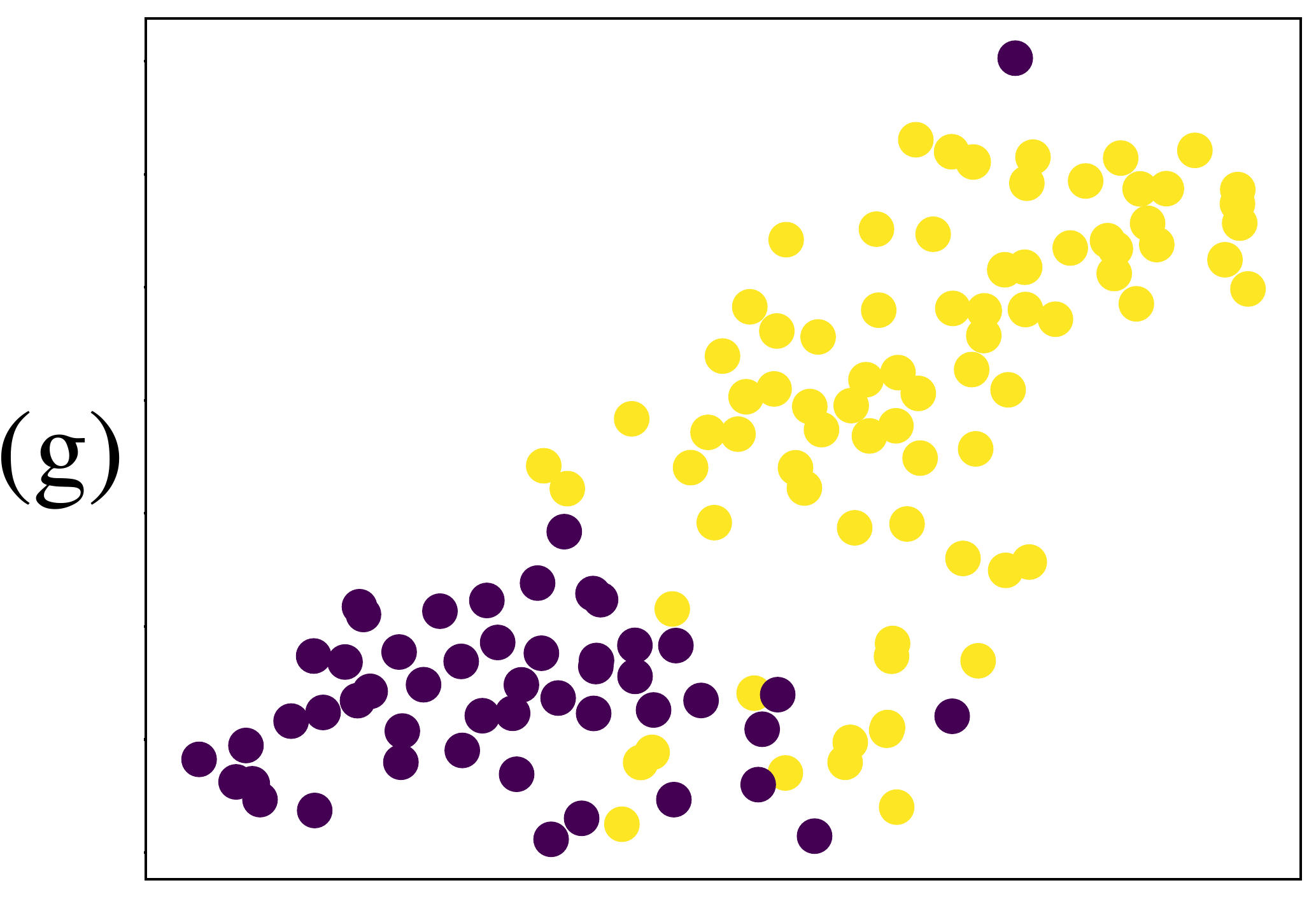}\end{center}
\end{minipage}
\caption{The visualization of some selected category masks and the visualization of the t-SNE feature embedding. We provide an example video of ``bending something so that it deforms'' and the corresponding different category masks of this input video: (a) ``bending something so that it deforms''; (b) ``bending something until it breaks''; (c) ``approaching something with your camera''; (d) ``closing something''; and (e) ``digging something out of something''. Moreover, the t-SNE feature embedding visualizations between two similar categories, ``bending something so that it deforms'' (yellow) and ``bending something until it breaks'' (purple), are shown in (f) and (g), which are without and with U2S feature embedding, respectively.}
\label{fig:category_mask}
\end{figure}

{Since masks are class-specific, we also provide an example of the learned video masks for different categories in Figure\ref{fig:category_mask} (a)-(e). This example reveals that different categories correspond to different regions. The two ``bending'' categories focus on where and when the object deforms or breaks. An interesting observation is that the category of ``approaching something with your camera'' mostly focuses on the background that may capture the camera movement. Moreover, in Figure\ref{fig:category_mask} (f) and (g), we also provide a visualization of the t-SNE figure embedding between two confusing categories, namely, ``bending something so that it deforms'' and ``bending something until it breaks''. The results show that the proposed U2S framework alleviates the similarity distributions of these confusing categories.}

\begin{table}[t]
\begin{center}
\vspace{-6pt}
\caption{Comparing the performance (top-1 classification accuracy (\%)) improvement by using the U2S framework between the confusing pairs shown in Fig.~\ref{fig:Mask}.}
\label{table:compare}
\begin{tabular}{cccc}
\toprule
Action                    & One-Pass & U2S   & Improved \\ \hline \hline
Bending sth. (deforms)     & 24.84    & 28.76 & 3.92     \\
Bending sth. (breaks)      & 66.99    & 71.84 & 4.85     \\ \hline \hline
Throwing sth. (pretending) & 38.30    & 48.94 & 10.64    \\
Throwing sth.             & 28.80    & 34.24 & 5.44     \\ \hline \hline
Putting sth. into sth.    & 45.21    & 47.60 & 2.39     \\
Taking sth. out of sth.   & 45.19    & 56.07 & 10.88    \\
Digging sth. out of sth.  & 54.55    & 60.61 & 6.06     \\
Pouring sth. out of sth.  & 63.29    & 75.95 & 12.66    \\ 
\bottomrule
\end{tabular}
\end{center}
\end{table}
\vspace{-6pt}

\subsubsection{Visualizing Weight Similarity Distributions}

\begin{figure*}[t]
\begin{center}
\subfloat[]{
      \includegraphics[width=0.25\linewidth]{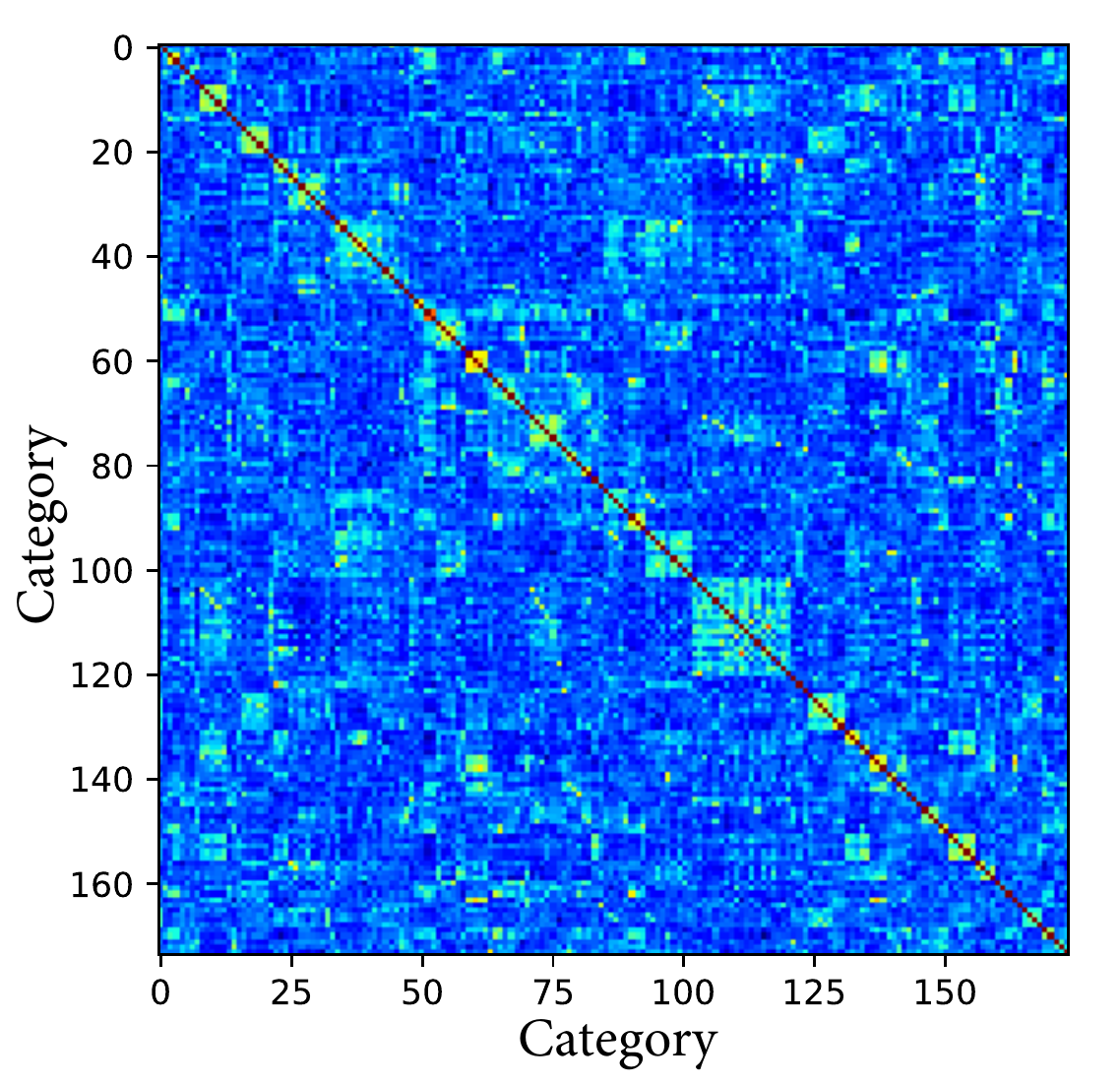}
}
\subfloat[]{
      \includegraphics[width=0.25\linewidth]{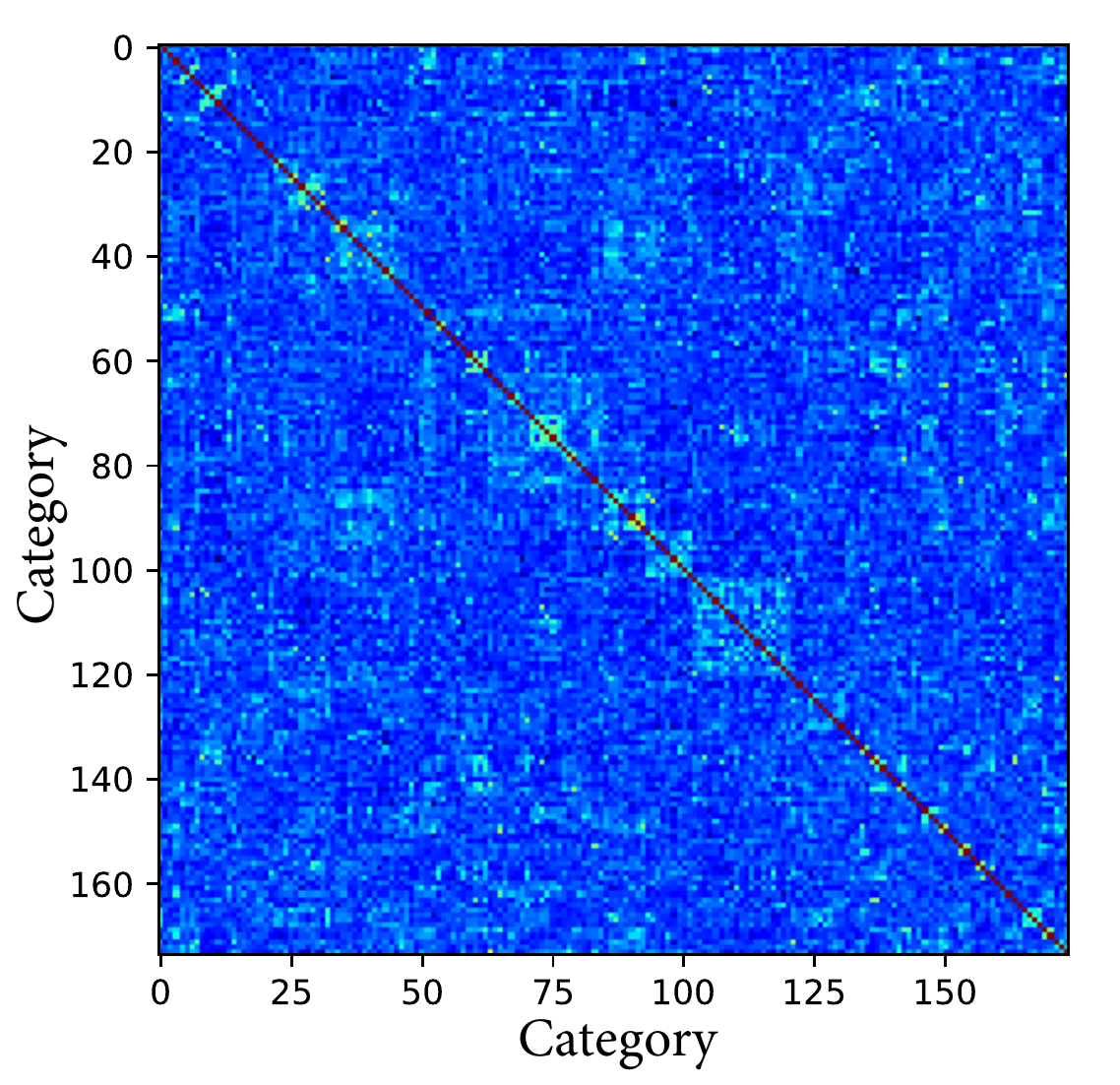}
}
\subfloat[]{
      \includegraphics[width=0.27\linewidth]{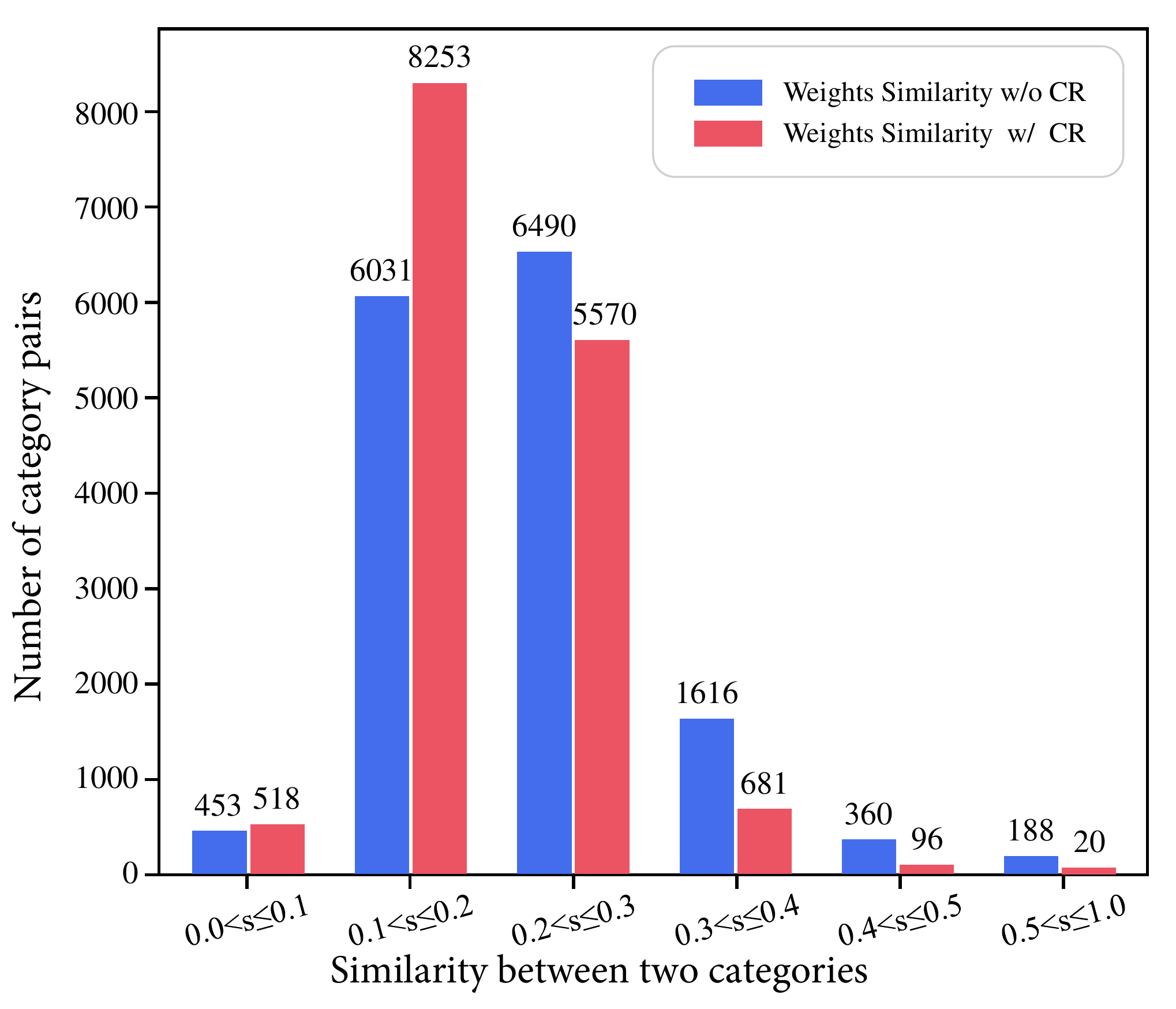}
}
\end{center}
\caption{Weight similarity distributions. (a) The weight similarity
distribution without category regularization. (b) The weight
similarity distribution with category regularization. (c) Histogram of the weight similarity. This figure is best viewed under magnification and in color.}
   \label{fig:weights}
\end{figure*}

In the ``Loss Function'' section, we introduced a category regularization method that restricts the discriminability between category-specific masks if these categories are easily confused. Therefore, in this section, we visualize the weight similarity among all categories in Fig.~\ref{fig:weights}. Similar to methods of visualizing feature similarity, the weights $\mathbf{W}\in\mathbb{R}^{C\times M}$ of the $conv 1\times 1\times 1$ layer that map the feature dimension to the number of categories are decomposed into category-wise weight vectors, $\mathbf{w}_{i}$. According to Eq.~\ref{eq:regularization1}, the weight similarity matrix $\mathbf{S}^{w}$ obtained by weight vectors can be visualized in Fig.~\ref{fig:weights}. The weight similarity distributions in (a) and (b) are without and with category regularization applied, respectively.
Clearly, the weight similarity without category regularization has a more cluttered distribution in the mask network than the weight similarity with category regularization.
$\mathbf{w}_{i}$ can be regarded as a classification boundary for category {\em i}.
Hence, classification boundaries between similar categories ({\em i.e.}, high values in $\mathbf{S}^{w}$) easily tend to be close without category regularization.
When category regularization is applied, some difficult category classification boundaries are constrained to relatively large distances. In the mask network, category regularization can help to learn category masks by a more discriminative feature combiner. Furthermore, categories are sorted by the first letter of the category name, so the same action verbs will be listed adjacent to one another. Thus, we can observe some square structures in the diagonal position. Comparing Fig.~\ref{fig:weights} (a) (b), some square structures disappear with the application of category regularization.
Therefore, adding category regularization to the mask network can help the network learn more category-specific masks.

Similar to Fig.~\ref{fig:feature} (b), we plot a weight similarity histogram to count the number of category pairs in different similarity intervals in Fig.~\ref{fig:weights} (c), which shows that more category pairs can become dissimilar when applying category regularization.

\begin{table*}[ht]
\begin{center}
\caption{Classification results on the Something-Something V2 dataset, which are compared with recently developed action recognition methods. $\ast$ means the result is based on our implementation.}
\label{table:results}
\begin{tabular}{cccccc}
\toprule
\multirow{2}{*}{Method}                                    & \multirow{2}{*}{Inputs} & \multicolumn{2}{c}{Validation} & \multicolumn{2}{c}{Test} \\
                                                            &                         & Top-1 (\%)      & Top-5 (\%)     & Top-1 (\%)   & Top-5 (\%)  \\ \hline
MultiScale TRN \cite{zhou2017temporal}                      & RGB                     & 48.80          & 77.64         & 50.85       & 79.33      \\
VGG style 11-layered 3D-CNN~\cite{goyal2017something}       & RGB                     & 51.33          & 80.46         & 50.76       & 80.77      \\
$\ast$ ECO$_{Lite}$ \cite{zolfaghari2018eco}                & RGB                     & 52.73          & 81.30         & -           & -          \\
$\ast$ TSM$_{8F}$ \cite{lin2018temporal}                    & RGB                     & 56.00          & 83.11         & -           & -          \\
$\ast$ TSM$_{8F}$ + U2S                                     & RGB                     & 57.40          & 83.73         & -           & -          \\
$\ast$TSM$_{16F}$ \cite{lin2018temporal}                    & RGB                     & 58.87          & 84.69         & -           & -          \\
$\ast$ TSM$_{16F}$ + U2S                                    & RGB                     & 60.13          & 85.34         & -           & -          \\
Our One-Pass                                                & RGB                     & 56.52          & 82.95         & -           & -          \\
Our U2S Model                                               & RGB                     & 58.47          & 84.03         & 58.01       & 83.71      \\ \hline \hline
2-Stream TRN \cite{zhou2017temporal}                        & RGB+flow                & 55.52          & 83.06         & 56.24       & 83.15      \\
Our One-Pass Model                                          & RGB+flow                & 59.37          & 85.48         & -           & -          \\ 
Our U2S Model                                               & RGB+flow                & 61.46          & 87.71         & 60.78       & 87.29      \\ 
\bottomrule
\end{tabular}
\end{center}
\end{table*}

\subsubsection{Comparison with State-of-the-Art Results}

We show the results of a comparison among state-of-the-art classification results on the Something-Something V2 dataset in Table~\ref{table:results}. The evaluations are performed on both the validation set and test set. Here, we discuss the results using RGB inputs on the validation dataset. The MultiScale TRN model \cite{zhou2017temporal} achieves a top-1 accuracy of 48.80\% and a top-5 accuracy of 77.64\%, which reflect the temporal relation between different frames. The baseline provided by~\cite{goyal2017something}, a simple VGG-style 11-layer 3D-CNN model, achieves a top-1 accuracy of 51.33\% and a top-5 accuracy of 80.46\%. Another 3D-CNN-based model, ECO$_{\mathrm{Lite}}$~\cite{zolfaghari2018eco}, which also combines 2D-CNN and 3D-CNN networks, achieves a top-1 accuracy of 52.73\% and a top-5 accuracy of 81.30\%. This lightweight structure is superior to the classic VGG-style 3D model. The TSM$_{16F}$ model \cite{lin2018temporal} achieves a top-1 accuracy of 58.70\% and a top-5 accuracy of 84.80\%, a state-of-the-art performance using only RGB inputs.
{Limited by computing resources, we also establish our U2S framework on the TSM$_{8F}$ and TSM$_{16F}$ models with a smaller batch size. As illustrated in Table~\ref{table:results}, the U2S framework can achieve a better performance on this backbone. }
Comparing the results of our framework with those of existing methods, U2S achieves comparable performance, namely, a top-1 accuracy of 58.47\% and a top-5 accuracy of 84.03\% on the validation set. The merged universal network and category-specific network can improve the accuracy by approximately 2\% over our one-pass model, which is already a strong baseline. This means that the U2S features benefit the final prediction. In addition, we feed only RGB image data to our network, and the proposed framework performs better than the 2-Stream TRN, which requires both RGB and optical flow inputs.

Considering the temporal relation network (TRN)~\cite{zhou2017temporal}, a large performance gain of 6.72\% is observed between the MultiScale TRN and 2-Stream TRN.
A 2.99\% performance gain is also obtained between our U2S model using only RGB or RGB+flow inputs. Thus, we can conclude that short motion information provided by optical flow helps greatly in such action recognition tasks.


\subsection{Results on the UCF101 and HMDB51 Datasets}

We provide more experimental results on the U2S framework in Table~\ref{table:ucf&hmdb} to confirm that our learning strategy is helpful on other action recognition datasets. We evaluate our approach on two common datasets, UCF101 and HMDB51. These datasets contain rich and diverse categories. Comparing the one-pass model and the U2S model, both flow and RGB data modality achieve a performance improvement of 2\% or 3\% on these two datasets. These experimental results are consistent with those of the Something-Something dataset. Similar to many action recognition tasks \cite{wu2015modeling, wang2016temporal, carreira2017quo}, the classification performance will improve when merging the RGB and optical flow results. We find large performance gains of 2.01\% and 8.5\% on the UCF101 and HMDB51 datasets, respectively.

\begin{table}[!t]
\begin{center}
\caption{Classification results on the UCF101 and HMDB51 datasets.}
\label{table:ucf&hmdb}
\begin{tabular}{ccccc}
\toprule
Dataset                 & Input                         & Method         & Top-1 (\%) & Top-5 (\%) \\ \hline
\multirow{5}{*}{UCF101} & \multirow{2}{*}{flow}         & One-Pass Model & 80.48    & 94.71    \\
                        &                               & U2S Model      & 82.66    & 95.08    \\ \cline{2-5} 
                        & \multirow{2}{*}{RGB}          & One-Pass Model & 89.16    & 98.57    \\
                        &                               & U2S Model      & 92.31    & 98.38    \\ \cline{2-5} 
                        & \multirow{2}{*}{RGB+flow}     & One-Pass Model & 92.43    & 98.73    \\ 
                        &                               & U2S Model      & \bf{94.32}    & \bf{98.89}    \\ \hline \hline
\multirow{5}{*}{HMDB51} & \multirow{2}{*}{flow}         & One-Pass Model & 53.40    & 80.13    \\
                        &                               & U2S Model      & 56.34    & 81.50    \\ \cline{2-5} 
                        & \multirow{2}{*}{RGB}          & One-Pass Model & 63.27    & 87.65    \\
                        &                               & U2S Model      & 65.49    & 89.80    \\ \cline{2-5} 
                        & \multirow{2}{*}{RGB+flow}     & One-Pass Model & 70.24    & 91.01    \\
                        &                               & U2S Model      & \bf{73.99}    & \bf{92.61}    \\ 
\bottomrule
\end{tabular}
\end{center}
\end{table}

\subsection{{Results on Image Datasets}}

\begin{figure}[t]
\begin{center}
      \includegraphics[width=0.85\linewidth]{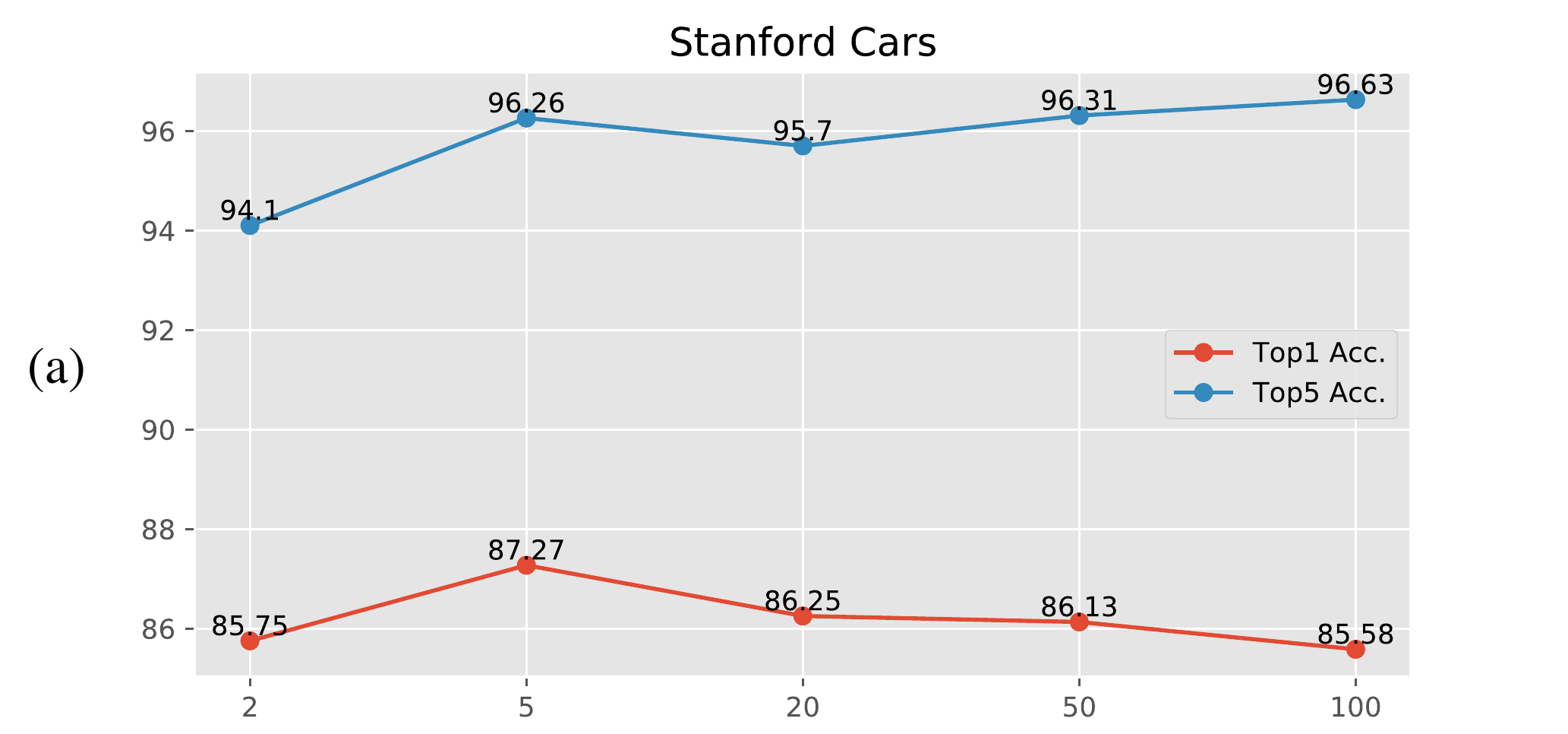}
\\
      \includegraphics[width=0.85\linewidth]{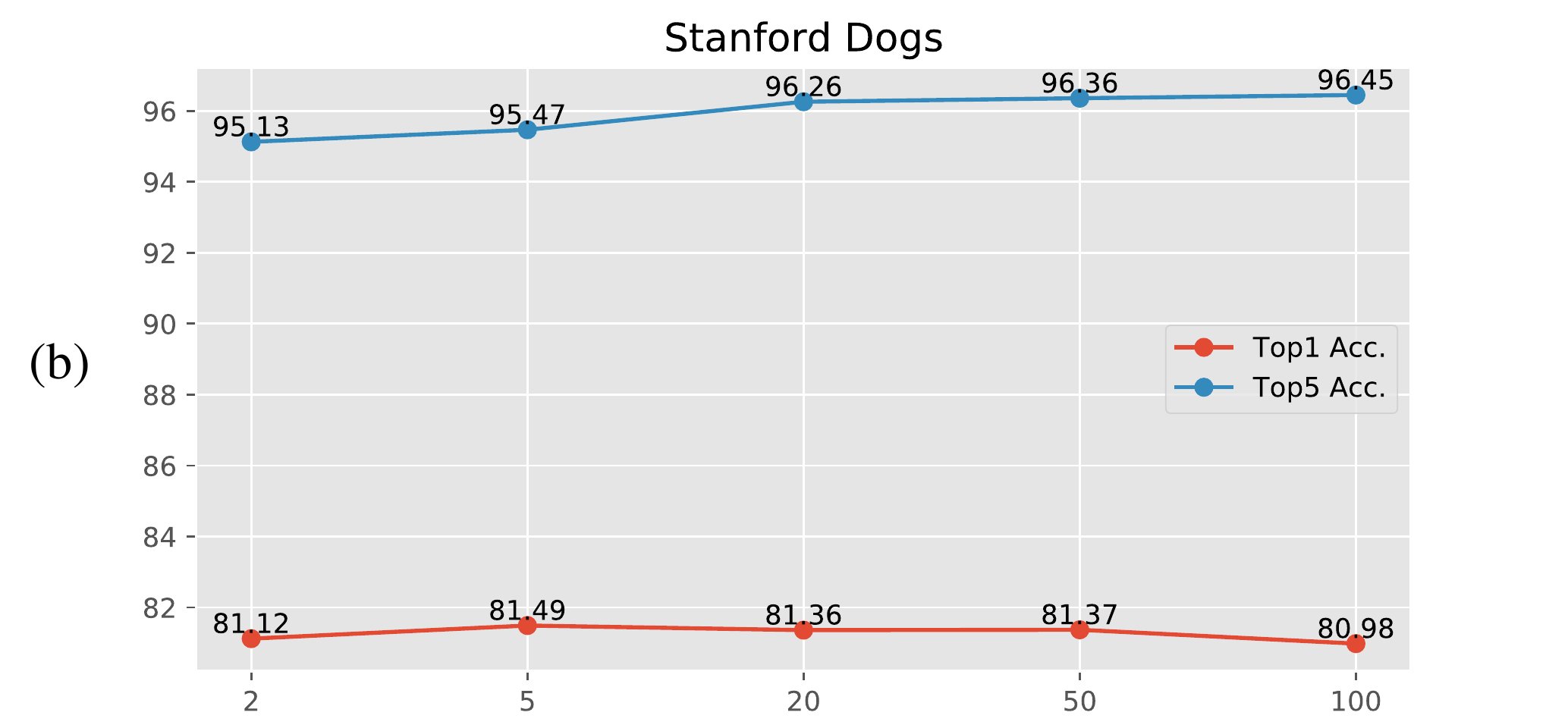}
\end{center}
\caption{The influence of the average number of confusing categories (N) on Stanford Cars (a) and Stanford Dogs (b).}
   \label{fig:threshold}
\end{figure}

{Since our proposed U2S framework is not limited to video data, it can also be applied to image classification tasks. We provide additional results on two fine-grained image datasets, Stanford Cars~\cite{KrauseStarkDengFei-Fei_3DRR2013} and Stanford Dogs~\cite{KhoslaYaoJayadevaprakashFeiFei_FGVC2011}, to show that the U2S framework is also effective for these fine-grained image classification tasks. As illustrated in Table~\ref{table:cars&dogs}, the top-1 accuracy is improved by an absolute 2.29\% for Stanford Cars and 6.09\% for Stanford Dogs, which are based on the ResNet50 backbone. Since these are fine-grained datasets and our method focuses on confusing categories, the top-5 accuracy also greatly improves. We further explore the influence of the average number of confusing categories (N) on the Stanford Cars/Dogs datasets in Fig.~\ref{fig:threshold}, which shows that the top-1 accuracy achieves a better result when N is set to 5, while the top-5 accuracy increases with increasing N.}

\begin{table}[t]
\begin{center}
\caption{Classification results on the Stanford Cars and Stanford Dogs datasets.}
\label{table:cars&dogs}
\begin{tabular}{cccc}
\toprule
Dataset                        & Method         & Top-1 (\%) & Top-5 (\%) \\ \hline
\multirow{2}{*}{Stanford Cars~\cite{KrauseStarkDengFei-Fei_3DRR2013}} & One-Pass Model & 84.98      & 93.97      \\
                               & U2S Model      & \textbf{87.27}      & \textbf{96.26}      \\ \hline
\multirow{2}{*}{Stanford Dogs~\cite{KhoslaYaoJayadevaprakashFeiFei_FGVC2011}} & One-Pass Model & 75.40      & 89.17      \\
                               & U2S Model      & \textbf{81.49}      & \textbf{95.47}      \\ 
\bottomrule
\end{tabular}
\end{center}
\end{table}

\section{Conclusions}

This paper proposes a U2S framework aimed at classifying complex fine-grained actions from the perspective of a ``rethinking'' mechanism to learn discriminative features. The U2S model is a generalized framework that employs universal features to guide category-specific features. The universal network and the category-specific network can be replaced by any base network and are suitable for many complex classification tasks.
We further introduce visualization methods based on feature similarity and weight similarity to evaluate the quality of the learned representations. Our experiments show that U2S indeed learns more discriminative features by visualizing feature correlations among different categories.

\appendices




\ifCLASSOPTIONcaptionsoff
\newpage
\fi




{
\small

\bibliographystyle{IEEEtran}
}




%

\begin{IEEEbiography}[{\includegraphics[width=1in,height=1.25in,clip,keepaspectratio]{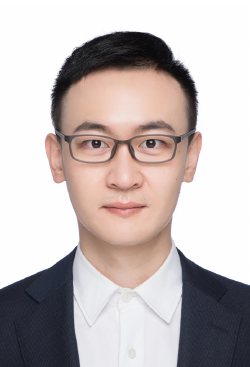}}]{Peisen Zhao}
received his B.E. degree from the University of Electronic Science and Technology of China (UESTC), Chengdu, China, in 2016. He has been working towards a Ph.D. at the Cooperative Meidianet Innovation Center, Shanghai Jiao Tong University, since 2016. His research interests include action recognition, action localization, and video analysis.
\end{IEEEbiography}

\vfill

\begin{IEEEbiography}[{\includegraphics[width=1in,height=1.25in,clip,keepaspectratio]{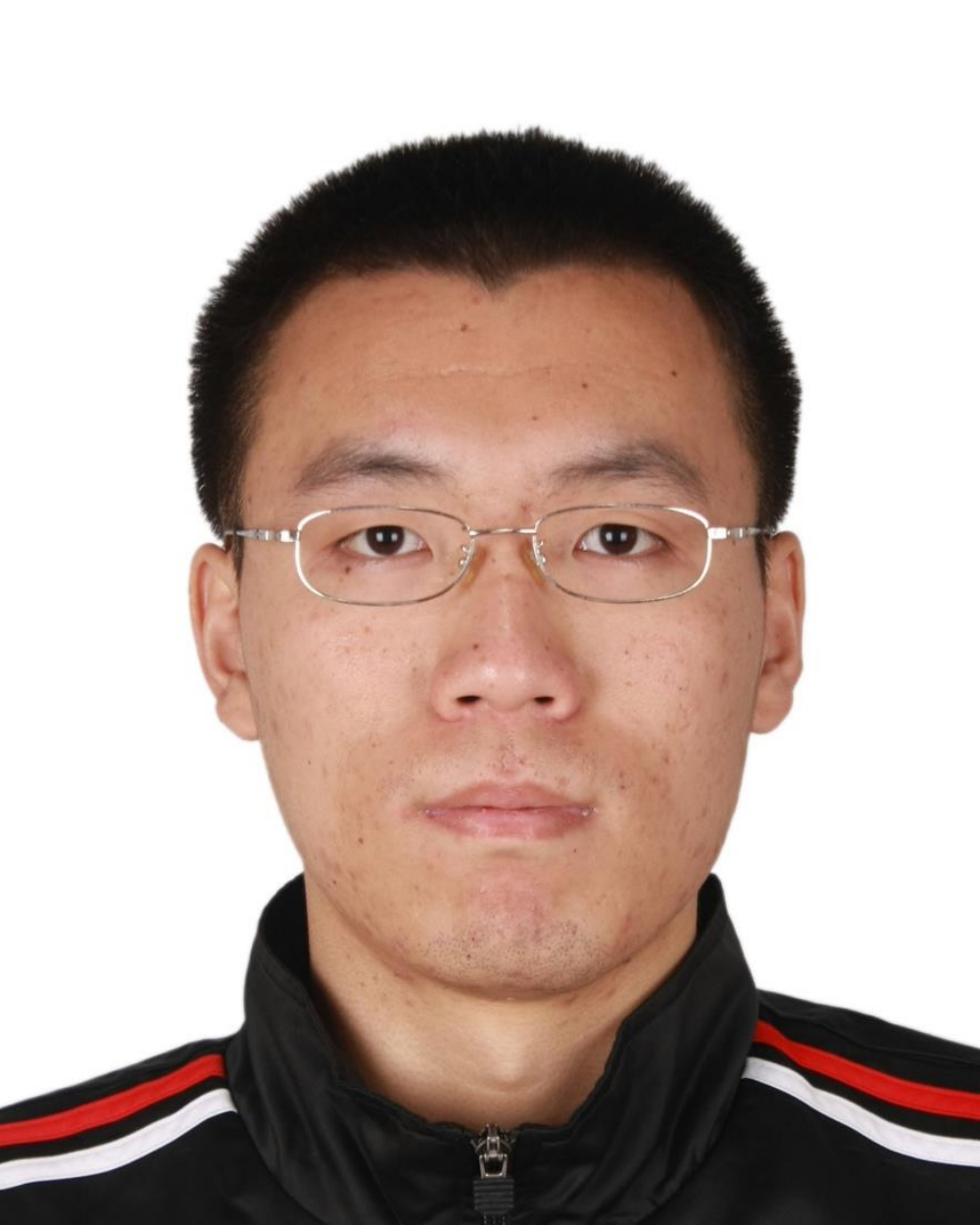}}]{Lingxi Xie}
is currently a senior researcher at Noah's Ark Lab, Huawei Inc. He received his B.E. and Ph.D. in engineering, both from Tsinghua University, in 2010 and 2015, respectively. He also served as a postdoctoral researcher at the CCVL lab from 2015 to 2019, having moved from the University of California, Los Angeles, to Johns Hopkins University.

Lingxi's research interests are in computer vision, particularly the application of deep learning models. His research covers image classification, object detection, semantic segmentation, and other vision tasks. He is also interested in medical image analysis, especially object segmentation in CT and MRI scans.

\end{IEEEbiography}

\vfill
\begin{IEEEbiography}[{\includegraphics[width=1in,height=1.25in,clip,keepaspectratio]{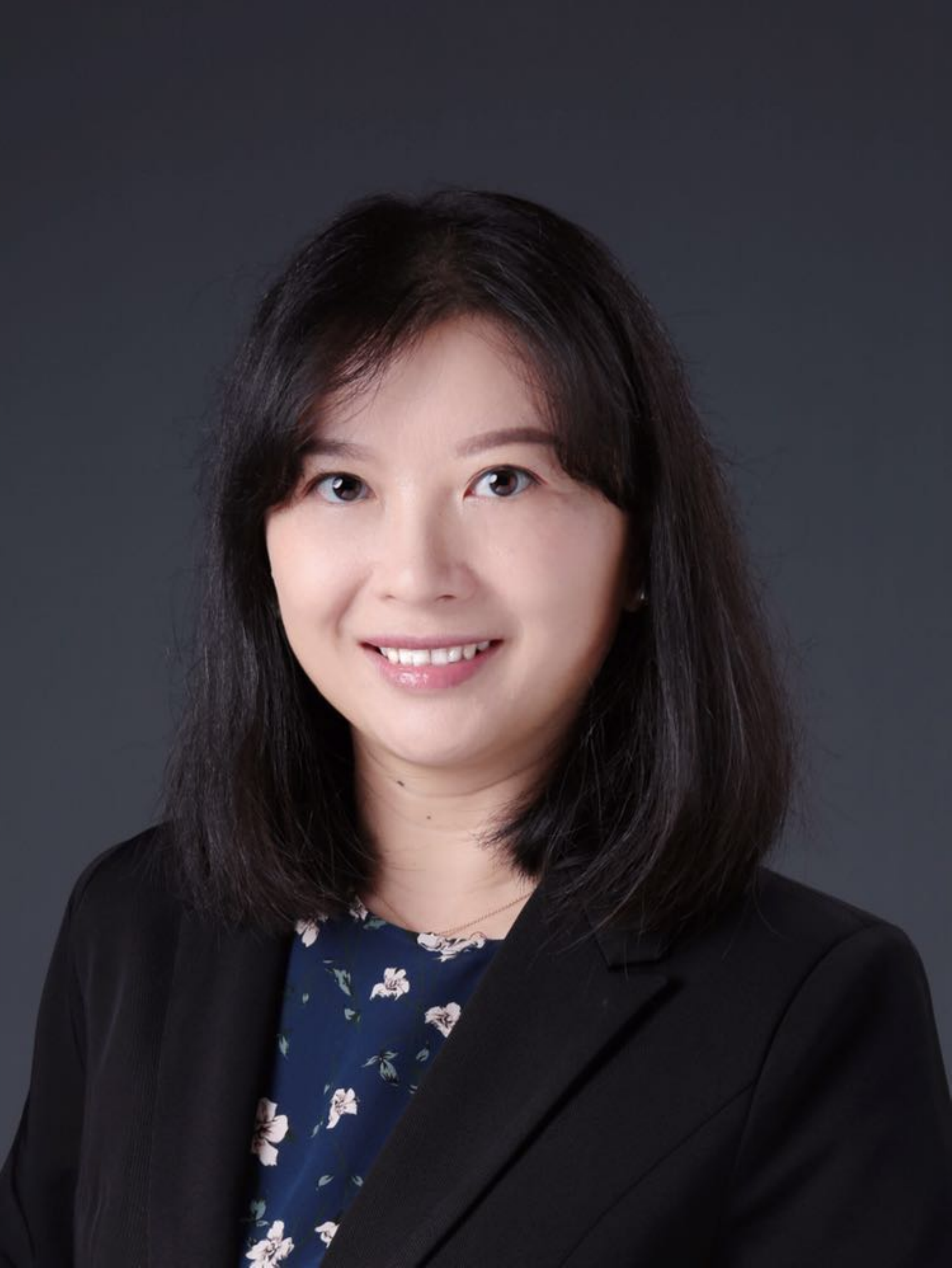}}]{Ya Zhang}
is currently a professor at the Cooperative Medianet Innovation Center, Shanghai Jiao Tong University. Her research interest is mainly in machine learning with applications to multimedia and healthcare. Dr. Zhang holds a Ph.D. degree in Information Sciences and Technology from Pennsylvania State University and a bachelor's degree from Tsinghua University in China. Before joining Shanghai Jiao Tong University, Dr. Zhang was a research manager at Yahoo! Labs, where she led an R\&D team of researchers with strong backgrounds in data mining and machine learning to improve the web search quality of Yahoo international markets. Prior to joining Yahoo, Dr. Zhang was an assistant professor at the University of Kansas with a research focus on machine learning applications in bioinformatics and information retrieval. Dr. Zhang has published more than 70 refereed papers in prestigious international conferences and journals, including TPAMI, TIP, TNNLS, ICDM, CVPR, ICCV, ECCV, and ECML. She currently holds 5 US patents and 4 Chinese patents and has 9 pending patents in the areas of multimedia analysis. She was appointed the Chief Expert for the 'Research of Key Technologies and Demonstration for Digital Media Self-organizing' project under the 863 program by the Ministry of Science and Technology of China.
\end{IEEEbiography}

\vfill
\begin{IEEEbiography}[{\includegraphics[width=1in,height=1.25in,clip,keepaspectratio]{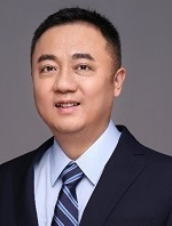}}]{Qi Tian}
is currently the Chief Scientist of computer vision at Huawei Noah's Ark Laboratory and a full professor with the Department of Computer Science at the University of Texas at San Antonio (UTSA). He was a tenured associate professor during 2008-2012 and a tenure-track assistant professor during 2002-2008. From 2008 to 2009, he took faculty leave for one year at Microsoft Research Asia (MSRA) as Lead Researcher in the Media Computing Group. Dr. Tian received his Ph.D. in ECE from the University of Illinois at Urbana-Champaign (UIUC) in 2002 and received his B.E. degree in electronic engineering from Tsinghua University in 1992 and his M.S. degree in ECE from Drexel University in 1996. Dr. Tian's research interests include multimedia information retrieval, computer vision, pattern recognition. He has published over 360 refereed journal and conference papers. He was the coauthor of a Best Paper at ACM ICMR 2015, a Best Paper at PCM 2013, a Best Paper at MMM 2013, a Best Paper at ACM ICIMCS 2012, a Top 10\% Paper at MMSP 2011, a Best Student Paper at ICASSP 2006, a Best Student Paper Candidate at ICME 2015, and a Best Paper Candidate at PCM 2007. Dr. Tian received the 2017 UTSA President's Distinguished Award for Research Achievement; the 2016 UTSA Innovation Award; the 2014 Research Achievement Awards from the College of Science, UTSA; the 2010 Google Faculty Award; and the 2010 ACM Service Award. He is an associate editor of many journals and on the Editorial Board of the Journal of Multimedia (JMM) and Journal of Machine Vision and Applications (MVA). He is a fellow of the IEEE.
\end{IEEEbiography}







\end{document}